\documentclass[10pt,5p]{elsarticle}

	\journal{Journal of Biomechanics}

\usepackage[T1]{fontenc}
\usepackage{stix2} 

	\usepackage{siunitx}  
	   					
	\usepackage{graphicx}
	\graphicspath{{figures/}{figures/introduction/}{figures/methods/}{figures/results/}{figures/discussion/}{figures/appendix/}}
	\usepackage[font=small]{caption} 
	\usepackage[font=normalsize]{subcaption} 
	
	\usepackage{tikz}
	\usetikzlibrary{positioning, shapes.multipart  }
	
	\tikzstyle{startstop} = [rectangle, rounded corners, minimum width=1cm, minimum height=1em,text centered, draw=black, fill=teal!20!white ]
	
	\tikzstyle{io} = [trapezium, trapezium left angle=70, trapezium right angle=110, minimum width=1cm, minimum height=1em, text width=2cm, text centered, draw=black, fill=black!10]
	
	\tikzstyle{process} = [rectangle, minimum width=1cm, minimum height=1em, text width=2.1cm, text centered, draw=black, fill=black!10]
	
	\tikzstyle{block} = [rectangle split, rectangle split parts=2, rounded corners, minimum width=3cm, minimum height=7em, text width=4.5 cm, text centered]
		
	\tikzstyle{decision} = [diamond, aspect=1.4, minimum width=0.5cm, minimum height=1em, text centered, text width=1.4cm, draw=black, fill=cyan!20!white]
	
	\tikzstyle{arrow} = [thick,->,>=stealth, darkblue]
	
	\usepackage{booktabs}

	\usepackage{amsmath,amsfonts}
	\usepackage{bm}

	\usepackage{xcolor}
 	
 	\definecolor{darkred}{rgb}{0.8500, 0.3250, 0.0980}  
 	\definecolor{mycyan}{rgb}{0.3010, 0.7450, 0.9330} 
 	\definecolor{darkblue}{rgb}{0 0.4470 0.7410}
 	\definecolor{darkyellow}{rgb}{0.9290 0.6940 0.1250} 	

	\usepackage{oplotsymbl} 	
 	\newcommand{\DeltaMarker}{{\scriptsize$\trianglepafill$}}
 	\newcommand{\SqMarker}{{\scriptsize$\squadfill$}}
	\newcommand{\CircMarker}{{\scriptsize$\circletfill$}}

	\DeclareRobustCommand{\tps}[3]{#1--#2--#3}  
	\DeclareRobustCommand{\fdps}[3]{#1=#2=#3} 
	\DeclareRobustCommand{\flex}[1]{$\tilde{\mathrm{#1}}$} 
	
	\newcommand{\imag}{\mathrm{i}}
	\newcommand{\euler}{\mathrm{e}}
		
	\usepackage[colorlinks=true,
				linkcolor=blue,
				citecolor=blue,
				urlcolor=blue,
				filecolor=blue,
				]{hyperref}

	\usepackage{lineno}
	\modulolinenumbers[1]

	\usepackage{natbib}
	\biboptions{authoryear}

\begin{document}

\begin{frontmatter}

	\title{Comprehending finger flexor tendon pulley system using  \\ a  computational analysis}

	\author{Vitthal Khatik
	}
	\ead{vitthal@iitk.ac.in}
	\author{Shyam Sunder Nishad
	}
	\ead{shyam@iitk.ac.in}
	\author{Anupam Saxena
		\corref{cor1}%
	}
	\ead{anupams@iitk.ac.in}
	\cortext[cor1]{Corresponding author}

	\address{Indian Institute of Technology Kanpur, Kanpur, India}

	\begin{abstract}
	Existing prosthetic/orthotic designs are rarely based on kinetostatics of a biological finger, especially its tendon-pulley system (TPS)  which helps render a set of extraordinary functionalities. Studies on computational models or cadaver experiments do exist. However, they provide little information on TPS configurations that lead to lower tendon tension, bowstringing, and pulley stresses, all of which a biological finger may be employing after all. A priori knowledge of such configurations and associated trade-offs is helpful not only from the design viewpoint of, say, an exoskeleton but also for surgical reconstruction procedures. We present a parametric study to determine optimal TPS configurations for the flexor mechanism. A compliant, flexure-based computational model is developed and simulated using the pseudo rigid body method, with various combinations of pulley/tendon attachment point locations, pulley heights, and widths. Deductions are drawn from the data collected to recommend the most suitable configuration. Many aspects of the biological TPS configuration are explained through the presented analysis. We reckon that the analytical approach herein will be useful in arriving at customized (optimized) hand exoskeletal designs.

	\end{abstract}
\end{frontmatter}


\section{Introduction}
Nature has gifted human fingers with the ability to perform extraordinarily diverse movements that are combinations of the four basic types- flexion, extension, and ad/abduction. Anatomically, flexor tendon pulley system (TPS) and extensor mechanism are responsible for transferring power from respective muscles to phalangeal bones, to perform these movements. In case of injury or post-stroke cognitive impairment, the patient may need TPS reconstruction surgery or artificial devices depending on cases and severity. Therefore, it is crucial to understand the biomechanics of the TPS involved.

Flexion-extension mechanics of a finger is studied mainly in two ways: by (i) performing surgical procedures on cadaver hands, and (ii) using computational models. Computational models provide a non-destructive alternative to studies using cadaver hands thereby reducing the number of cadaver surgeries required. Given the complex anatomy of the hand, developing an accurate and generic model is still being actively researched and remains a challenging task.

\begin{figure}[h!]
	\centering
	\includegraphics[scale=1]{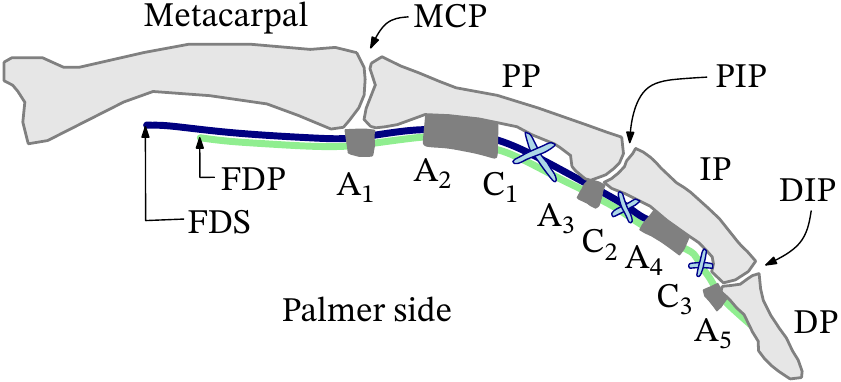}
	\caption{ Schematic of biological flexor tendon-pulley system (TPS) in a human finger: It comprises of two tendons (FDP and FDS) with five annular pulleys (A$_1$-A$_5$) and three cruciate (cross-shaped) pulleys (C$_1$-C$_3$). Pulleys are fibrous tissue bands that keep tendons close to finger bones. Annular pulleys are stiff, while cruciate pulleys are flexible but inextensible. Pulleys A$_2$ and A$_4$ are the most important for full range flexion (\cite{dy2013flexor, chow2014importance}). PP -- proximal phalange, IP -- intermediate phalange, DP -- distal phalange. MCP -- metacarpophalangeal joint, PIP -- proximal interphalangeal joint, DIP -- distal interphalangeal joint. FDP -- flexor digitorum profundus, FDS -- flexor digitorum superficialis.  }
	\label{fig:tendon_pulley}
\end{figure}

Several studies exist on the role of each pulley and tendon in the flexor TPS (Fig. \ref{fig:tendon_pulley}).  Among annular pulleys, A$_2$ is the widest, followed by A$_4$. Stress in pulley fibers depend on pulley location and width. High pulley stress can cause discomfort in finger movement  (\cite{schweizer2008biomechanics}). Some cadaver studies suggest that smaller width of A$_2$ and A$_4$ pulleys can be used without compromising much on the flexion range (\cite{mitsionis1999feasibility, chow2014importance, leeflang2014role}). Those by \cite{kauko1967positioning} and \cite{hume1991biomechanics} that employ computational models contradict on the effect of change in pulley positions.  Loosening/removal of pulleys can cause bowstringing (i.e., tendon moving away from bones during flexion-extension), which reduces the range of flexion (ROF), creating difficulty in forming a fist, or grasping (\cite{dy2013flexor,brand1975tendon}). While designing artificial systems or  performing TPS reconstruction surgery,  proper knowledge of pulley locations, widths, and heights (i.e., loosening),  and, exclusion of pulleys/tendon if required and the trade-offs involved, will help make the best possible decisions. 

Existing literature does not focus much on the tendon tension requirement for finger flexion. It is desirable to have the highest range of flexion (ROF) for a given tendon tension for the corresponding muscle load to be reduced. This is even more critical when designing bionic artificial devices, as, selection of actuators, and battery power requirements  would pose a limit on this tension. This paper presents a parametric study to arrive at optimal flexor TPS configurations, which maximizes ROF while keeping bowstringing and pulley stress as small as possible. We also investigate whether having both FDS and FDP tendons improves the flexion range, or one can be ignored to simplify the design of an artificial device without significant loss in functionality.

The paper is organized as follows. The computational biomechanical model developed is presented in section \ref{sec:methods}. Parametric study and results are described in section \ref{sec:results}. TPS configurations are recommended and comparisons with the biological finger made in section \ref{sec:discussion}, followed by conclusion in section \ref{sec:conclusion}.

\section{Methods}
\label{sec:methods}

\begin{figure}[t!]
	\centering
	\includegraphics[scale=1]{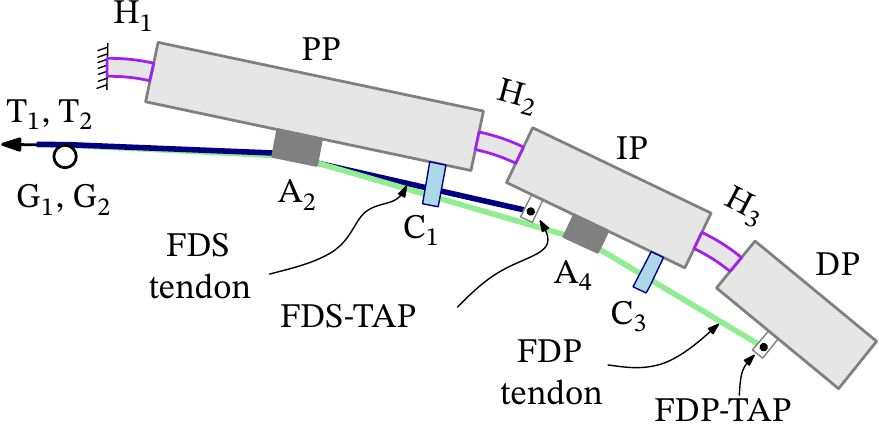}
	\caption{Suggested tendon-pulley system model for finger flexion: Phalange bones modeled as stiff beams of uniform cross-sections. Essential pulleys A$_2$, A$_4$, C$_1$, and C$_3$ retained from Fig. \ref{fig:tendon_pulley}. C--pulleys keep bowstringing low and interact with a tendon only when taut (Fig. \ref{fig:loose_Cpulley}).   Interphalangeal joints modeled using flexure hinges (H$_1$, H$_2$, and H$_3$). Inextensible strings of negligible thickness, with tensions $T_1$ and $T_2$, used for the FDP and FDS tendons passing over ground pulleys G$_1$ and G$_2$, and attaching at FDP--TAP and FDS--TAP, respectively. TAP stands for tendon attachment point.}	
	\label{fig:TPS_model}
	\vspace{1.5em}
	
	\includegraphics[scale=1]{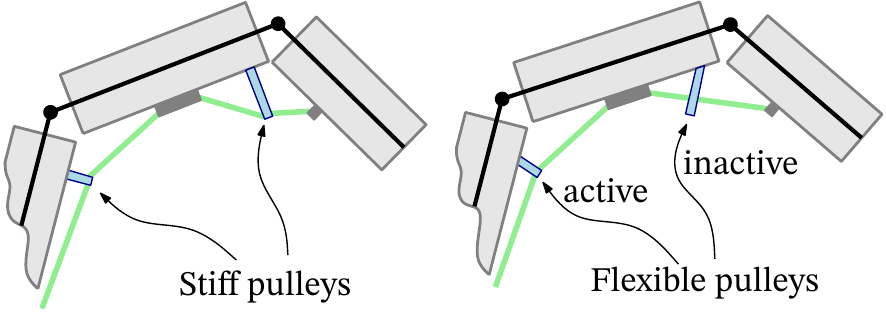}
	\caption{Flexible--inextensible C--pulleys compared with stiff pulleys: A flexible C-pulley  reorients itself to be free of bending moment from tendon tension. It remains inactive unless pulled by any tendon.}
	\label{fig:loose_Cpulley}
 
\end{figure}

We modeled the flexor mechanism of an index finger\footnote{Index finger is the most dexterous among all four fingers, and is, therefore, a good choice for framing a generic model.} as a beam-string arrangement described in Fig. \ref{fig:TPS_model}. Human finger joints have non-fixed axes of rotation and inherent stiffnesses (\cite{van2008natural}). Hence, we modeled them as flexure hinges which possess similar deformation characteristics (\cite{guo2013compliant}). Biological finger tendons experience minimal strain (\cite{pring1985mechanical}), and thus, were modeled as inextensible strings. C--pulleys are cross-shaped (cruciate), flexible--inextensible, and remain loose unless
pulled by a tendon. We implemented these characteristics through flexible--inextensible string loops (Fig. \ref{fig:loose_Cpulley}). 

\begin{figure*}[h!]
	\centering
	\includegraphics[scale=1]{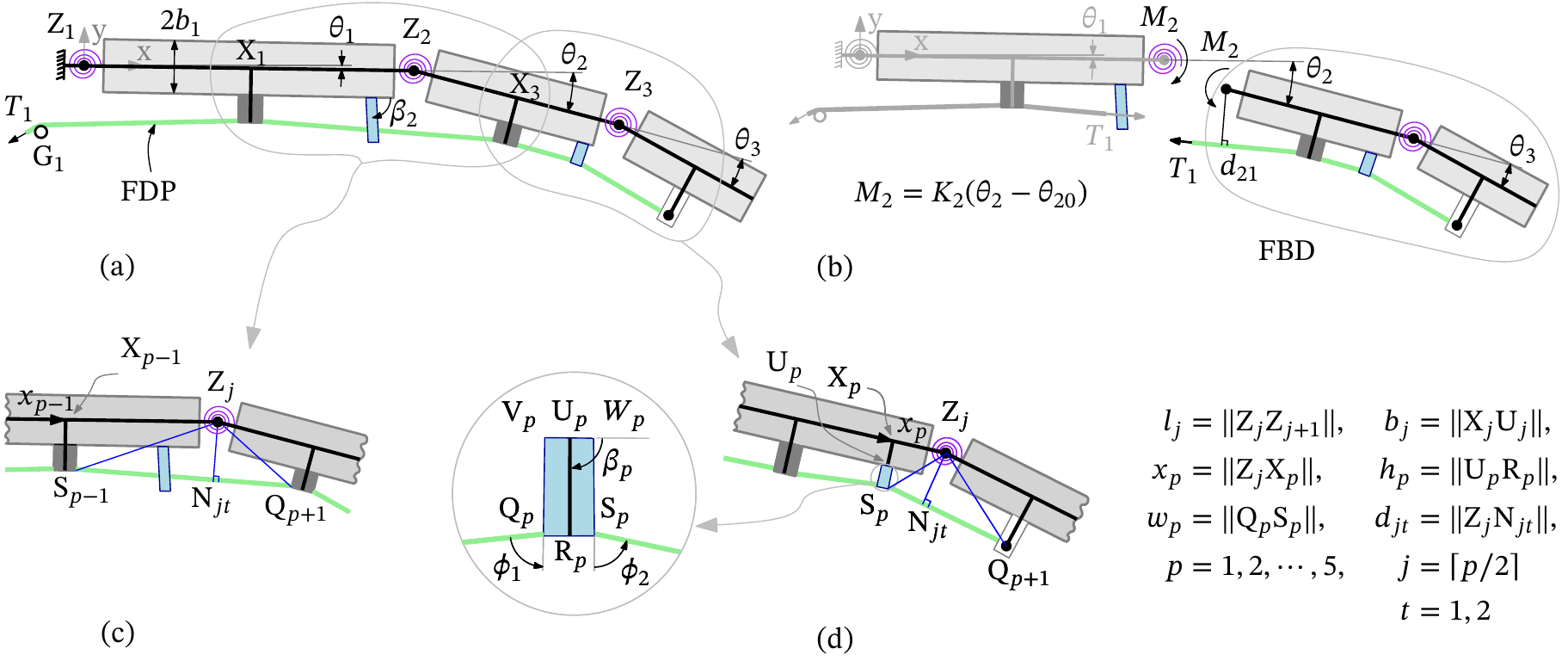}
	\caption{ Equivalent 3R Pseudo-Rigid-Body Model (PRBM) shown in (a): Each flexure in its undeformed state is replaced by a revolute joint at its center (Z$_j$) and a torsional spring. Moment balance to solve the joint kinetics: Free body diagram (FBD) shown in (b) of the distal portion of finger attached to torsional spring at the PIP joint. Assumed external force absent. $K_2$ is the stiffness and $\theta_{20}$ is the neutral position of torsional spring equivalent to the PIP joint.
   Computation of moment arm at $j^\mathrm{th}$ joint with inactive $p^\mathrm{th}$ pulley shown in (c) and active $p^\mathrm{th}$ pulley in (d): Pulley proximal to the joint, shown as an example. All vertices of $\Delta$Z$_j$S$_\xi$Q$_{p+1}$ ($\xi = p$ or $p-1$), and therefore moment arm $d_{jt}$ of the tension $T_t$ can be expressed in terms of joint angles $\theta_j$, pulley locations $x_p$, heights $h_p$, and widths $w_p$. $2b_j$ is the bone width of the $j^\text{th}$ phalange. $t=1$, and $2$ denote FDP and FDS tendons, respectively. The pulley angle $\beta_p$ is $-90^{\circ}$ for a stiff pulley, and needs to be computed for a flexible-inextensible C--pulley. }	
  \label{fig:prbm}	
\end{figure*}

\subsection{Mathematical Formulation} 
For full range flexion, the flexure hinges must undergo large bending deflection with small strain. A 2R or 3R pseudo-rigid-body model (PRBM) of such a flexure is computationally simpler than the nonlinear finite element method (FEM), and gives much smaller approximation error compared to a 1R PRBM (\cite{su20093R_PRBM} and \cite{yue20122R_PRBM}). Nevertheless, we chose 1R PRBM because (i) in our analysis, it gave results sufficiently close to those from FEM and the experiments performed (\ref{app:fem} and \ref{app:validation}), and (ii) this study concerns only with relative responses of different TPS configurations.  The proposed model of the overall TPS is thus, 3R, as shown in Fig. \ref{fig:prbm}a. We assumed that the finger moves slowly while changing its posture. Hence, joint velocities and accelerations were considered zero. All external (contact and non-contact) forces were assumed absent. Friction was neglected in tendon-pulley contacts.  With these assumptions, governing equations of the quasistatic system can be written from Fig. \ref{fig:prbm}b as:
\begin{align}
	\label{eq:moment_full}
	K_j (\theta_j - \theta_{j0}) = \sum_{t=1}^{2} T_t\, d_{jt} , \quad j=1,2,3, \quad d_{32} = 0  
\end{align}
where, $K_j$ is the stiffness and $\theta_{j0}$ is the neutral position of the torsional spring at  $j^\mathrm{th}$ joint. Here, $j = 1,2, 3$ identify the three joints MCP, PIP, and DIP (and the three phalanges PP, IP, and DP), respectively. $\theta_j$ is the corresponding joint angle.  Moment arm $d_{jt}$ at the $j^\mathrm{th}$ joint is $\|\mathrm{Z}_j \mathrm{N}_{jt}\|$ (\cite{kauko1967positioning}) for  $t^\mathrm{th}$ tendon tension $T_t$, as shown in Fig. \ref{fig:prbm}c--d. Indices $t = 1, 2$ correspond to  FDP and FDS tendons, respectively. Since FDS tendon does not exert moment on the DIP joint, the moment arm $d_{32}$ was set to zero. Moment arm $d_{jt}$ can be computed in terms of the TPS parameters, as follows. 

Let $p = 0, 1,2, \cdots$ denote the sequence of pulleys/TAPs in which they are connected with a given tendon. Ground pulley is always indexed as 0. As an example, $p = 1$ and $2$ for pulleys C$_1$ and A$_2$ , respectively, if  pulley C$_1$ is proximal to pulley A$_2$. This indexing is performed independently for the two tendons.  Let $x_{p}$, $w_{p}$, and $h_{p}$ be relative position, width, and height of $p^\mathrm{th}$ pulley (Fig. \ref{fig:prbm}). The pulley end is  marked with points Q$_p$, R$_p$ (midpoint), and S$_p$ (Fig. \ref{fig:prbm}d) occupying positions $q_p, r_p,$ and $s_p$, respectively. Let $j^\mathrm{th}$ joint (Z$_j$) occupy position $z_j$. Using complex algebra, we may write:
\begin{equation}
	\label{eq:vertices}
	\left.
	\begin{aligned}
		z_1 &= 0, \, z_2  = l_1 \euler^{\imag \theta_1}, \, z_3 = z_2 + l_2 \euler^{\imag (\theta_1 + \theta_2) } 
		\\
		r_p &= z_j + \left[x_p - \imag b_j + h_p \euler^{\imag\beta_p} \right] \euler^{\imag \left(\sum_{k=1}^{j} \theta_k\right) } 
		\\
		q_p &= r_p - \frac{w_p}{2} \euler^{\imag\left(\beta_p+\pi/2 + \sum_{k=1}^{j} \theta_k \right) }  
		\\
		s_p &= r_p + \frac{w_p}{2} \euler^{\imag\left(\beta_p+\pi/2 + \sum_{k=1}^{j} \theta_k \right) }  \\
		\text{if $p^\mathrm{th}$ } &\text{pulley is on $j^\mathrm{th}$ phalange } 
	\end{aligned}   
	\right\}
\end{equation}
Here, $l_j$ and $2b_j$ are length and nominal bone-width of $j^\mathrm{th}$ phalange, respectively.
We assumed $\beta_p = -\pi/2$ for stiff pulleys. If the flexible-inextensible C--pulley is active, as in Figs. \ref{fig:loose_Cpulley} and  \ref{fig:prbm}d, it orients itself along the angle bisector of the two segments of tendon in contact. This ensures zero bending moment in that C--pulley. Hence, the correponding angle $\beta_p$ is solved by minimizing the following objective:
\begin{equation}
	\begin{aligned} 
		\Delta\phi = |\phi_1 - \phi_2| 
	\end{aligned} 
\end{equation}
where, $\phi_1$ and $\phi_2$ are angles with the pulley direction (U$_\mathrm{p}$R$_\mathrm{p}$) made by the left and right tendon segments (Q$_p$S$_{p-1}$) and (S$_p$Q$_{p+1}$), respectively, as shown in Fig. \ref{fig:prbm}d (enlarged view). 
For a flexible pulley indexed $p$, $\phi_1$ and $\phi_2$ can be expressed in terms of pulley parameters and joint angles as follows:
\begin{align}
	\phi_1 &= \arg \frac{r_{p} - u_{p}}{s_{p-1} - q_{p}},   &
	\phi_2 &= \arg \frac{q_{p+1} - s_{p}}{r_{p} - u_{p}}
\end{align}
$p^\mathrm{th}$ pulley becomes active when the shortest distance of the tendon segment   S$_{p-1}$Q$_{p+1}$ from the base U$_p$ becomes equal to or smaller than the pulley height. In that case, the moment arm base N$_{jt}$ lies on S$_{p}$Q$_{p+1}$, otherwise on S$_{p-1}$Q$_{p+1}$.  In case flexible pulley is located distally relative to the joint, N$_{jt}$ lies on S$_{p-1}$Q$_{p}$ when the pulley is active, and S$_{p-1}$Q$_{p+1}$ otherwise.
\begin{figure*}
	\centering
	\begin{tikzpicture}
	[node distance=6mm and 6mm]
	\node[block, rectangle split part fill={magenta!50, magenta!20}, draw=magenta] (increment) { \textbf{Increment tensions}
		\nodepart{two}
		$T_\mathrm{s} = T_\mathrm{s} + \delta T_\mathrm{s}$ until 
		$
		\begin{aligned} 
			T_1 \, \text{or}\, T_2 & \ge T_\mathrm{m},\, \text{where,} \\
			T_1 &= (1-\gamma) \, T_\mathrm{s} \\
			T_2 &= \gamma \, T_\mathrm{s}  
		\end{aligned}
		$
		};
	\node[block, right = of increment, rectangle split part  fill={cyan!50, cyan!20}, draw=cyan] (solve) 
	{ 	\textbf{Solve for free joints}
		\nodepart{two}
		Solve $\theta_j$ from Eq. \eqref{eq:moment_full} \\[5pt]
		Find all $j=i \,\big|\, \theta_i > \theta_{i\mathrm{m}}$ \\[5pt]
		For each $i$, solve $T_\mathrm{s}\, \text{and} \, \theta_{j\ne i}$
		with $\theta_i = \theta_{i\mathrm{m}}$.
    };
	
	\node[block, right = of solve, rectangle split part  fill={yellow!50, yellow!20}, draw=yellow] (freeze) 
	{
		\textbf{Joint-Locking}
		\nodepart{two} 
		Find $k=i \, \big|\, \displaystyle  \min_i  T_\mathrm{s} $ \\[2pt]
        Record $\theta_{j\ne k}$, $\theta_{k\mathrm{m}}$, and $\displaystyle \min_i T_\mathrm{s}$ 
         \\[5pt]
        Drop $k^\mathrm{th}$ Eq. from \eqref{eq:moment_full}
		 
	};

	\draw[arrow] (increment) -- (solve);
	\draw[arrow] (solve) -- (freeze);
	\draw[arrow] (freeze) -- ++(0,-2cm) node(){}    -|  (increment);
	\end{tikzpicture}
	\caption{Procedure for solving Eq. \eqref{eq:moment_full} involving joint-locking and two tendons. $T_\mathrm{s}$ is the sum $T_1 + T_2$ of tensions in FDP and FDS tendons. $\gamma$ is the ratio $T_2/T_\mathrm{s}$ which we choose a priori when both tendons are actuated. Subscript m indicates maximum permissible values. Iterations continue until maximum tension is reached or all joints lock. Owing to discrete nature of numerical simulations, $\theta_j$ can exceed the limit $\theta_{j\mathrm{m}}$ in an incremental step, for one or more joints.} 
	\label{fig:algo}
\end{figure*}
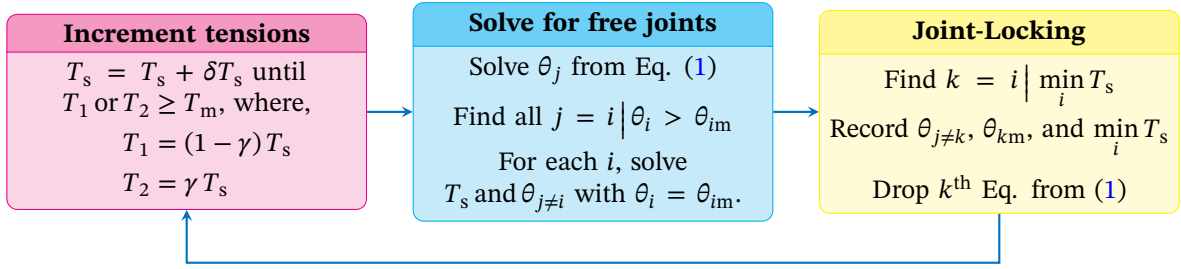

Point N$_{jt}$ ($n_{jt}$) on the nearest segment S$_\xi$Q$_\eta$ ($\xi = p$ or $p-1$, $\eta= p$ or $p+1$) can be found using the following two conditions:\\
(i) Z$_j$N$_{jt}$ $\perp$ S$_{\xi}$Q$_{\eta}$ implying:
\begin{align} 
(n_{jt} - z_j ) (s_{\xi}-q_{\eta})^* + (n_{jt}-z_j)^* (s_{\xi}-q_{\eta} ) = 0 
\end{align} 
(ii) N$_{jt}$ lies on the line S$_{\xi}$Q$_{\eta}$. Therefore,
\begin{align}
\label{eq:alpha} 
n_{jt} = (1- \alpha) s_{\xi} + \alpha q_{\eta}
\end{align} 
where $\alpha \in \mathbb{R} $. Solving these two equations yields:
\begin{align}
n_{jt} &= \frac{z_j}{2} + \frac{1}{2} \frac{ s_{\xi}^* \,\, q_{\eta} - q_{\eta}^* \,\, s_{\xi} }{ s_{\xi}^* - q_{\eta}^* } + \frac{z_j^*}{2} \frac{s_{\xi} - q_{\eta}}{s_{\xi}^*- q_{\eta}^*} \\
d_{jt} &= | z_j - n_{jt} |
\end{align}

To obtain flexion response of the TPS, we incremented tensions $T_1$ and T$_2$ in a given ratio $\gamma$, and solved Eqs. \eqref{eq:moment_full} for joint angles at each step (Fig. \ref{fig:algo}). For this, we employed the trust-region-dogleg optimization algorithm via the fsolve() implementation of MATLAB. We also simulated the joint rotation limits of a biological finger during flexion. Physically,  interphalangeal contact invalidates the equation corresponding to the locked joint in the set \eqref{eq:moment_full}. Therefore, we dropped that equation and solved  others for the remaining unknown joint angles and the tendon tensions.  In case two or more joints locked in the same $T$-step,  we found the joint that locked at the lowest tension in that incremental step, and recorded only the corresponding state of the TPS.  

\subsection{Finger flexion, Bowstringing, and Pulley stress}

We quantified flexion by the sum $\Sigma\,\theta_j$. For two different TPS configurations, identical values of $\Sigma\,\theta_j$ need not imply that the joints angles are also identical. Still, $\Sigma\,\theta_j$ indicates the amount of finger curl and is, therefore, a useful quantity to compare TPS configurations.
Bowstringing at a joint was quantified as the shortest distance between the joint and a tendon. For instance, bowstringing $B_{21}$ at PIP joint ($j=2$) due to FDP tendon ($t=1$), when C$_1$ pulley ($p=2$) is active, is:
\begin{align}
	B_{21} =
	\begin{cases}
		\mathrm{\|Z_2N_{21}\|} = d_{21}  &\text{if $0 < \alpha < 1$}
		\\
		\mathrm{\|Z_2S_2\|} &\text{if $\alpha \le$ 0}
		\\
		\mathrm{\|Z_2Q_3\|} &\text{if $\alpha \ge$ 1}
	\end{cases} 
	\\
	\mathrm{N}_{21} \text{ on } \mathrm{S}_2\mathrm{Q}_3, \text{ see Eq. \eqref{eq:alpha} for } \alpha 
	\nonumber 
\end{align}
We defined $\max_{j, t} {B}_{jt}$ as the critical value $B_\mathrm{w}$.
We computed axial and bending stresses on a pulley (the enlarged view, Fig. \ref{fig:prbm}d) due to FDP and FDS tendons as follows: 
\begin{equation}
	\left.
	\begin{aligned}
		\sigma_\mathrm{axial} &= \frac{1}{wD} (T_1+T_2)(\cos\phi_1 + \cos\phi_2) \\
		\sigma_\mathrm{bending} &= h (T_1+T_2) (\sin\phi_1 - \sin\phi_2) \frac{ w/2 }{I} \\
		\sigma_\mathrm{net} &= |\sigma_\mathrm{axial}| + |\sigma_\mathrm{bending}| 
	\end{aligned}
	\right\}
	\label{eq:PS}
\end{equation}
where, $h$, $w$, $D$ and $I= Dw^{3} / 12 $ are the height, width, depth and area moment of inertia of the pulley. The maximum resultant stress exists at either V$_p$ or W$_p$ on the pulley base (Fig. \ref{fig:prbm}d), where bending stress is maximum. When computing stresses in flexible-inelastic pulleys, the pulley-width is assumed to remain $w$ even when pulley bends, and the pulley tip-line QS normal to the pulley length RU. This works well for pulleys of small widths. Biologically, FDS tendon terminates just proximally to the A$_4$ pulley. Likewise, it may not interact with pulleys A$_4$ and C$_3$ if they are located distally to FDS-TAP in a candidate TPS configuration. In that case, we substitute $T_2$ = 0 for them in Eqs. \eqref{eq:PS}. We defined the highest $\sigma_\mathrm{net}$ among all pulleys as the critical value $PS$.

\section{Results}
\label{sec:results}

\begin{table*}  
	\centering
	\caption{Finger Model Data: Joint stiffnesses $K_1$, $K_2$, and $K_3$ in \SI{}{\newton\milli\metre\per\deg}, as per \cite{kim2019development, dionysian2005proximal, kamper2002extrinsic}. Phalange lengths $l_1$, $l_2$, and $l_3$ in \SI{}{\milli\meter}, measured joint-to-joint, inclusive of flexure length $l_\mathrm{f}$ (Fig. \ref{fig:prbm}). Default values of pulley widths $w_0$, heights $h_0$, out-of-plane thicknesses $t_0$, bone-widths $b_0$, and offsets $e_0$ (Fig. \ref{fig:fdp_convention})  in \SI{}{\milli\meter}. Bone-width assumed equal for all phalanges. Ground pulleys G$_1$ and G$_2$ assumed coincident at ($X_\mathrm{g}, Y_\mathrm{g}$). Tendon tension   incremented in $n$ steps upto $T_0$ \SI{}{\newton}. Neutral positions $\theta_{10}, \theta_{20}$, and $\theta_{30}$ of finger joints, assumed to coincide with fully extended state, to simulate full range of motion as flexion.  Finger joint limits $\theta_{1\mathrm{m}}, \theta_{2\mathrm{m}}$, and $\theta_{3\mathrm{m}}$, as per \cite{zheng2010kinematics}. } 
	\label{tab:constants}
\begin{tabular}{c c p{8mm} c c p{8mm} c c  p{8mm} c c  }
	\hline
	\hline
	\addlinespace[3pt]
	Constants & Values 		& & Constants	& Values  & & Constants & Values 		& & Constants	& Values  \\
	\addlinespace[3pt]
	\hline
	\addlinespace[3pt]
	$K_1$ 	& 0.95 			& & $l_1$ 		& 42.0  & & $\theta_{10}$ 	& 0 			& & $\theta_{1\mathrm{m}}$		& 90$^\circ$  \\
	$K_2$ 	& 0.60 			& & $l_2$ 		& 27.0  & & $\theta_{20}$ 	& 0 			& & $\theta_{2\mathrm{m}}$ 		& 100$^\circ$  \\
	$K_3$ 	& 0.60 			& & $l_3$ 		& 19.5  & & $\theta_{30}$ 	& 0 			& & $\theta_{3\mathrm{m}}$ 		& 80$^\circ$  \\
	\addlinespace[3pt]
	\hline
	\addlinespace[3pt]
	$h_0$ 	& 0.5  			& & $w_0$ 		& 1.0 & & $X_\mathrm{g}$ 	& $-7.5$ & & $Y_\mathrm{g}$  & 	$-5.0$  \\
	$t_0$ 	& 10.0   		& & $b_0$ 		& 7.0 & & $T_0$ 	& 8.0      		& & $n$		 	& 200 \\
	$l_\mathrm{f}$ & 5.0    & & $e_0$       & 4.0 & & & & & & \\
	\addlinespace[3pt]
	\hline
\end{tabular}
	\vspace{1em}
	\caption{ Range of Flexion (ROF) and critical values of bowstringing ($B_\mathrm{w}$), and pulley stress (PS) for FDP-TPS configurations yielding ROF $>240^\circ$ and FDS-TPS configurations yielding ROF $>150^\circ$. Subscripts to $B_\mathrm{w}$ and PS values indicate the joints and pulleys, respectively, where those critical values occur. Table arranged in descending order of ROF at 8 N tension in FDP or FDS tendon individually. Superscript $\dagger$ implies  ROF, $B_\mathrm{w}$, and $PS$ values at 6.4 N tension in the FDP and FDS tendons when both tendons are actuated simultaneously with equal tensions. Offsets $e_j$, of locations P and D, of 10\% distance from joints (marked with superscript 10) calculated as percentage of bone lengths as:  ${e_j = ^{10}}e_{j} = (l_j - l_\mathrm{f})/10 + l_\mathrm{f}/2$, where $j=1,2,3$ indicates PP, IP, and DP, respectively. Heights $h_a$ and widths $w_a$ of pulleys A$_2$ and A$_4$. Similarly, $h_c$ and $w_c$ are heights and widths of pulleys C$_1$ and C$_3$. Heights of FDP-TAP and FDS-TAP $h_0$ in all cases. In some cases, MCP joint was ignored while computing B$_\mathrm{w}$ (Figs. \ref{fig:fdp_C_loose_height}, \ref{fig:fdp_C_location}, \ref{fig:fdp_height}). }
	\label{tab:all_results}
	\vspace{1em}
	\begin{tabular}{ c c c S[table-format=2.1, table-number-alignment = left, table-space-text-post
			=\textsubscript{MCP}   ] S[table-format=2.1, table-number-alignment = left, , table-space-text-post
			=\textsubscript{A4}   ] c c c c c c }
	\hline
	\hline
	\addlinespace[3pt]
	\multicolumn{2}{c}{ROF at} & TPS & $B_\mathrm{w}$ & $PS$ & $e_j$ & $h_a$ & $h_c$ & $w_a$ & $w_c$ & Refer to marked   \\	
	5 N & 8 N &  Configuration & {(mm)}  & {(MPa)} & (mm) & (mm) & (mm) & (mm) & (mm) &  curves in figures
	\\ 
	\addlinespace[3pt] 
	\hline
	\addlinespace[3pt]
251$^\circ$ & {$^\dagger$}270$^\circ$ & \fdps{C\flex{D}}{C\flex{D}}{C} & {$^\dagger$}9.0$_\text{ MCP}$ & {$^\dagger$}2.8$_\text{ C1}$ & {$^{10}e_j$} & $h_0$ & 2.0 & 2.0 & $w_0$ & {\color{darkblue} \CircMarker}, Fig. \ref{fig:fdps} \\ 
252$^\circ$ & {$^\dagger$}270$^\circ$ & \fdps{C\flex{D}}{C\flex{D}}{D} & {$^\dagger$}9.0$_\text{ MCP}$ & {$^\dagger$}3.0$_\text{ C1}$ & {$^{10}e_j$} & $h_0$ & 2.0 & 2.0 & $w_0$ & {\color{darkred} \CircMarker}, Fig. \ref{fig:fdps} \\ 
176$^\circ$ & 256$^\circ$ & \tps{C\flex{D}}{C\flex{D}}{C} & 8.9$_\text{ PIP}$ & 1.8$_\text{ C1}$ & {$^{10}e_j$} & $h_0$ & 2.0 & $w_0$ & $w_0$ & {\color{darkblue} \SqMarker}, Fig. \ref{fig:fdp_C_location}, {\color{darkblue} \DeltaMarker}, Fig. \ref{fig:fdp_height} \\ 
176$^\circ$ & 256$^\circ$ & \tps{C\flex{D}}{C\flex{D}}{C} & 9.0$_\text{ MCP}$ & 2.2$_\text{ C1}$ & {$^{10}e_j$} & $h_0$ & 2.0 & 2.0 & $w_0$ & {\color{darkblue} \DeltaMarker}, Fig. \ref{fig:fdps} \\ 
176$^\circ$ & 256$^\circ$ & \tps{C\flex{D}}{C\flex{D}}{D} & 9.0$_\text{ MCP}$ & 2.2$_\text{ C1}$ & {$^{10}e_j$} & $h_0$ & 2.0 & 2.0 & $w_0$ & {\color{darkred} \DeltaMarker}, Fig. \ref{fig:fdps} \\ 
200$^\circ$ & 256$^\circ$ & \tps{C\flex{D}}{C\flex{D}}{C} & 9.8$_\text{ PIP}$ & 1.8$_\text{ C1}$ & $e_0$ & $h_0$ & 4.5 & $w_0$ & $w_0$ & {\color{darkblue} \CircMarker}, Fig. \ref{fig:fdp_C_loose_height} \\ 
200$^\circ$ & 256$^\circ$ & \tps{C}{C}{C} & 10.3$_\text{ PIP}$ & 0.2$_\text{ A4}$ & -- & $h_0$ & -- & 8.0 & -- & {\color{mycyan} \CircMarker}, Fig. \ref{fig:fdp_width} \\ 
223$^\circ$ & 256$^\circ$ & \tps{C}{C}{C} & 13.1$_\text{ PIP}$ & 0.9$_\text{ A4}$ & -- & $h_0$ & -- & 2.0 & -- & {\color{mycyan} \SqMarker}, Fig. \ref{fig:fdp_width} \\ 
223$^\circ$ & 256$^\circ$ & \tps{C}{C}{D} & 13.6$_\text{ PIP}$ & 2.2$_\text{ A4}$ & $e_0$ & $h_0$ & -- & $w_0$ & -- & {\color{black} \CircMarker}, Fig. \ref{fig:fdp_tension}b, \ref{fig:fdp_Bw}b, \ref{fig:fdp_PS}b \\ 
223$^\circ$ & 256$^\circ$ & \tps{C}{C}{C} & 13.6$_\text{ PIP}$ & 2.4$_\text{ A4}$ & -- & $h_0$ & -- & $w_0$ & -- & {\color{black} \SqMarker}, Fig. \ref{fig:fdp_tension}b, \ref{fig:fdp_Bw}b, \ref{fig:fdp_PS}b \\ 
223$^\circ$ & 256$^\circ$ & \tps{C}{C}{C} & 13.8$_\text{ PIP}$ & 7.5$_\text{ A4}$ & -- & $h_0$ & -- & 0.5 & -- & {\color{mycyan} \DeltaMarker}, Fig. \ref{fig:fdp_width} \\ 
224$^\circ$ & 255$^\circ$ & \tps{C}{C}{C} & 14.1$_\text{ PIP}$ & 7.3$_\text{ A4}$ & -- & 2.0 & -- & $w_0$ & -- & {\color{mycyan} \SqMarker}, Fig. \ref{fig:fdp_height} \\ 
226$^\circ$ & 254$^\circ$ & \tps{C}{C}{C} & 14.8$_\text{ PIP}$ & 13.7$_\text{ A4}$ & -- & 3.5 & -- & $w_0$ & -- & {\color{mycyan} \CircMarker}, Fig. \ref{fig:fdp_height} \\ 
177$^\circ$ & 252$^\circ$ & \tps{C\flex{D}}{C\flex{D}}{C} & 8.9$_\text{ PIP}$ & 2.1$_\text{ C1}$ & {$^{10}e_j$} & 2.0 & 2.0 & $w_0$ & $w_0$ & {\color{darkblue} \SqMarker}, Fig. \ref{fig:fdp_height} \\ 
200$^\circ$ & 250$^\circ$ & \tps{C\flex{P}}{C\flex{D}}{C} & 13.6$_\text{ PIP}$ & 2.4$_\text{ A4}$ & {$^{10}e_j$} & $h_0$ & 2.0 & $w_0$ & $w_0$ & -- \\ 
148$^\circ$ & 247$^\circ$ & \tps{C\flex{D}}{C\flex{D}}{C} & 9.0$_\text{ MCP}$ & 1.4$_\text{ C1}$ & {$^{10}e_j$} & $h_0$ & $h_0$ & 8.0 & $w_0$ & {\color{darkblue} \CircMarker}, Fig. \ref{fig:fdp_width} \\ 
166$^\circ$ & 246$^\circ$ & \tps{CD}{CD}{C} & 7.9$_\text{ PIP}$ & 8.1$_\text{ C1}$ & {$^{10}e_j$} & $h_0$ & 2.0 & $w_0$ & $w_0$ & {\color{darkyellow} \SqMarker}, Fig. \ref{fig:fdp_C_location} \\ 
176$^\circ$ & 246$^\circ$ & \tps{C\flex{D}}{C\flex{D}}{C} & 8.9$_\text{ PIP}$ & 2.2$_\text{ C1}$ & {$^{10}e_j$} & 3.5 & 2.0 & $w_0$ & $w_0$ & {\color{darkblue} \CircMarker}, Fig. \ref{fig:fdp_height} \\ 
184$^\circ$ & 246$^\circ$ & \tps{C}{D}{P} & 18.9$_\text{ PIP}$ & 1.8$_\text{ A2}$ & $e_0$ & $h_0$ & -- & $w_0$ & -- & {\color{mycyan} \DeltaMarker}, Fig. \ref{fig:fdp_tension}b, \ref{fig:fdp_Bw}b, \ref{fig:fdp_PS}b \\ 
151$^\circ$ & 245$^\circ$ & \tps{C\flex{D}}{C\flex{D}}{C} & 9.0$_\text{ MCP}$ & 1.5$_\text{ C1}$ & {$^{10}e_j$} & $h_0$ & $h_0$ & 2.0 & $w_0$ & {\color{darkblue} \SqMarker}, Fig. \ref{fig:fdp_width} \\ 
151$^\circ$ & 245$^\circ$ & \tps{C\flex{D}}{C\flex{D}}{C} & 9.0$_\text{ MCP}$ & 1.6$_\text{ C1}$ & {$^{10}e_j$} & $h_0$ & $h_0$ & 0.5 & $w_0$ & {\color{darkblue} \DeltaMarker}, Fig. \ref{fig:fdp_width} \\ 
187$^\circ$ & 243$^\circ$ & \tps{C}{D}{C} & 18.9$_\text{ PIP}$ & 1.8$_\text{ A2}$ & $e_0$ & $h_0$ & -- & $w_0$ & -- & {\color{mycyan} \SqMarker}, Fig. \ref{fig:fdp_tension}b, \ref{fig:fdp_Bw}b, \ref{fig:fdp_PS}b \\ 
187$^\circ$ & 242$^\circ$ & \tps{C}{C}{P} & 13.6$_\text{ PIP}$ & 2.4$_\text{ A4}$ & $e_0$ & $h_0$ & -- & $w_0$ & -- & {\color{black} \DeltaMarker}, Fig. \ref{fig:fdp_tension}b, \ref{fig:fdp_Bw}b, \ref{fig:fdp_PS}b \\ 
187$^\circ$ & 242$^\circ$ & \tps{C}{D}{D} & 18.9$_\text{ PIP}$ & 1.8$_\text{ A4}$ & $e_0$ & $h_0$ & -- & $w_0$ & -- & {\color{mycyan} \CircMarker}, Fig. \ref{fig:fdp_tension}b, \ref{fig:fdp_Bw}b, \ref{fig:fdp_PS}b \\ 
114$^\circ$ & 176$^\circ$ & \tps{C\flex{D}}{C}{} & 9.0$_\text{ MCP}$ & 2.2$_\text{ C1}$ & {$^{10}e_j$} & $h_0$ & 2.0 & 2.0 & $w_0$ & {\color{darkblue} \SqMarker}, Fig. \ref{fig:fdps} \\ 
143$^\circ$ & 176$^\circ$ & \tps{C}{C}{} & 13.6$_\text{ PIP}$ & 0.8$_\text{ A2}$ & -- & $h_0$ & -- & $w_0$ & -- & {\color{black} \SqMarker}, Fig. \ref{fig:fds_all}b, \ref{fig:fds_all}c, \ref{fig:fds_all}d  \\ 
143$^\circ$ & 176$^\circ$ & \tps{C}{D}{} & 18.9$_\text{ PIP}$ & 1.8$_\text{ A2}$ & $e_0$ & $h_0$ & -- & $w_0$ & -- & {\color{black} \CircMarker}, Fig. \ref{fig:fds_all}b, \ref{fig:fds_all}c, \ref{fig:fds_all}d  \\ 
114$^\circ$ & 174$^\circ$ & \tps{C\flex{D}}{D}{} & 9.0$_\text{ MCP}$ & 2.8$_\text{ C1}$ & {$^{10}e_j$} & $h_0$ & 2.0 & 2.0 & $w_0$ & {\color{darkred} \SqMarker}, Fig. \ref{fig:fdps} \\ 
	\addlinespace[3pt]
	\hline
\end{tabular}
\end{table*}

\begin{figure}[]
	\centering
	\includegraphics[scale=1]{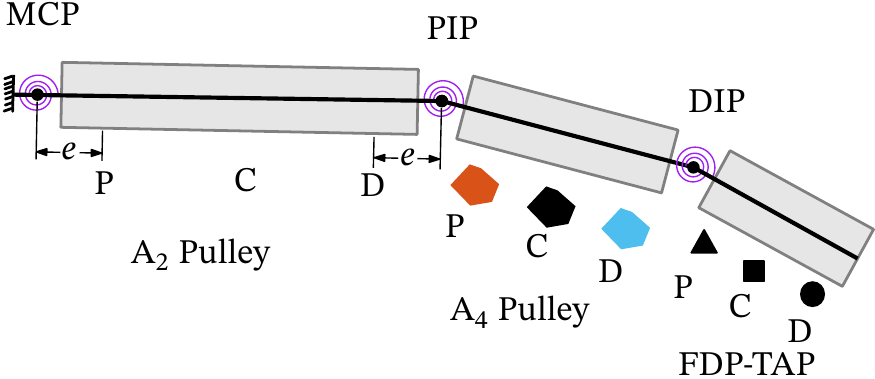}
	\caption{Three locations -- proximal (P), central (C) and distal (D) considered for each pulley or FDP--TAP. Numerically, P stands for $x = e$, C for $x = l/2$, and D for $x = l - e$, where $l$ is the corresponding phalange length (refer Fig. \ref{fig:prbm} for notation $x$). Default value of the offset $e=e_0=4$ mm. Note: In Figs. \ref{fig:fdp_tension}-\ref{fig:fdp_PS}, nine configurations corresponding to identical A$_2$ pulley location, shown in a single subplot. Colors and markers differentiate locations of A$_4$ pulley and the FDP--TAP, respectively. For example, the red delta marked curves in  the subplots (a) of Figs. \ref{fig:fdp_tension}--\ref{fig:fdp_PS} correspond to the FDS--TPS configuration in which  A$_2$ pulley, A$_4$ pulley and FDP--TAP are located proximally on their respective phalanges.    }	
	\label{fig:fdp_convention}
\end{figure}

\begin{figure*}[h!]
	\centering 
	\includegraphics[scale=0.795]{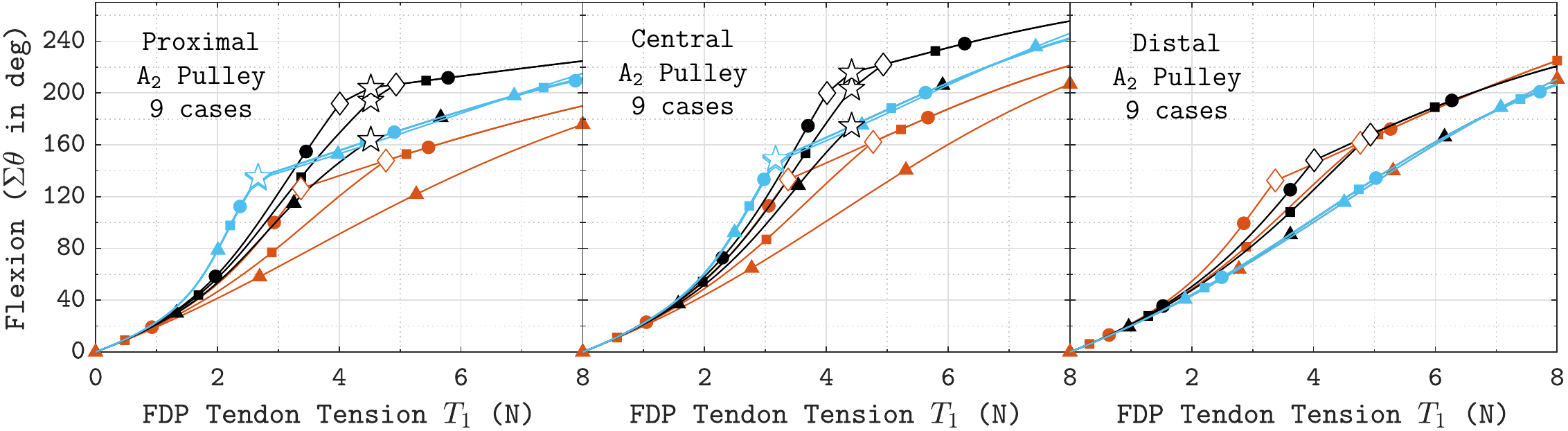} 
	\caption{a-c, left-right: FDP tendon tension-flexion response: Locations of pulley A$_4$ (denoted by different colors) and FDP--TAP (denoted by different markers) change in each subplot which corresponds to a single location of pulley A$_2$. Refer to Fig. \ref{fig:fdp_convention} for details on colors and marker symbols. Kinks on curves marked with star and diamond indicate PIP and DIP joint locking, respectively. All pulleys/FDP--TAP heights $h_0$, widths $w_0$, out of plane thicknesses $t_0$, and $a = a_0$. Higher the curve, better is the TPS configuration. 	
	}  
	\label{fig:fdp_tension} 
	\vspace{1em}   
	\centering 
	\includegraphics[scale=0.785]{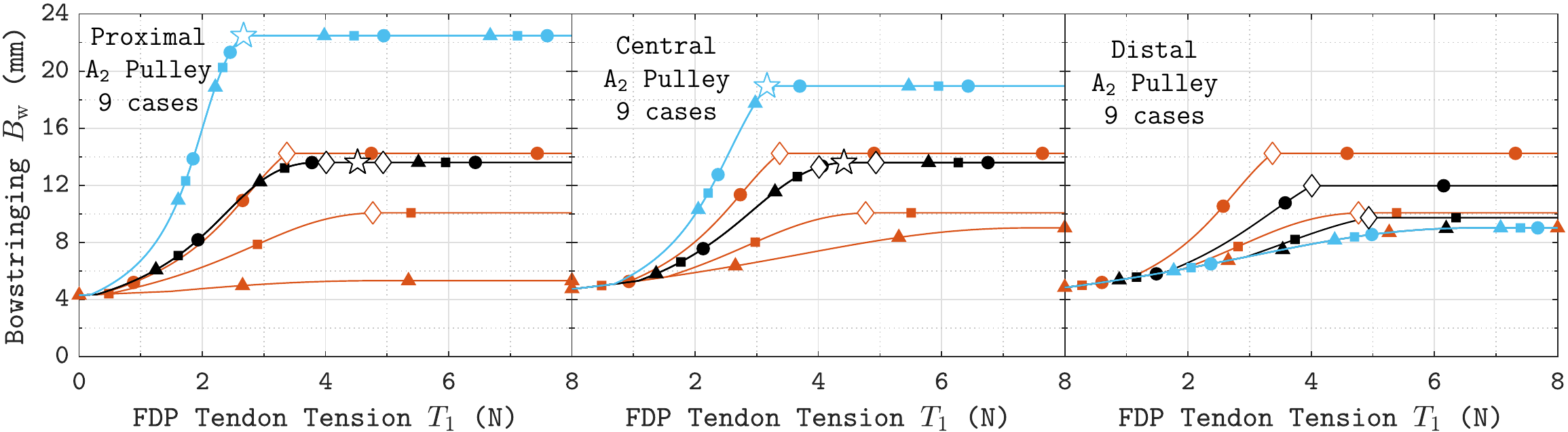} 
	\caption{a-c, left-right: FDP Bowstringing response: Curves arranged similar to Fig. \ref{fig:fdp_tension} and correspond to the same parameter values therein. Bowstringing is maximum near a joint. The largest of the three values near each joint was chosen as the critical value of bowstringing (Bw).  Lower the curve, better is the TPS configuration.
	}  
	\label{fig:fdp_Bw} 
	\vspace{1em}   
	\centering 
	\includegraphics[scale=0.785]{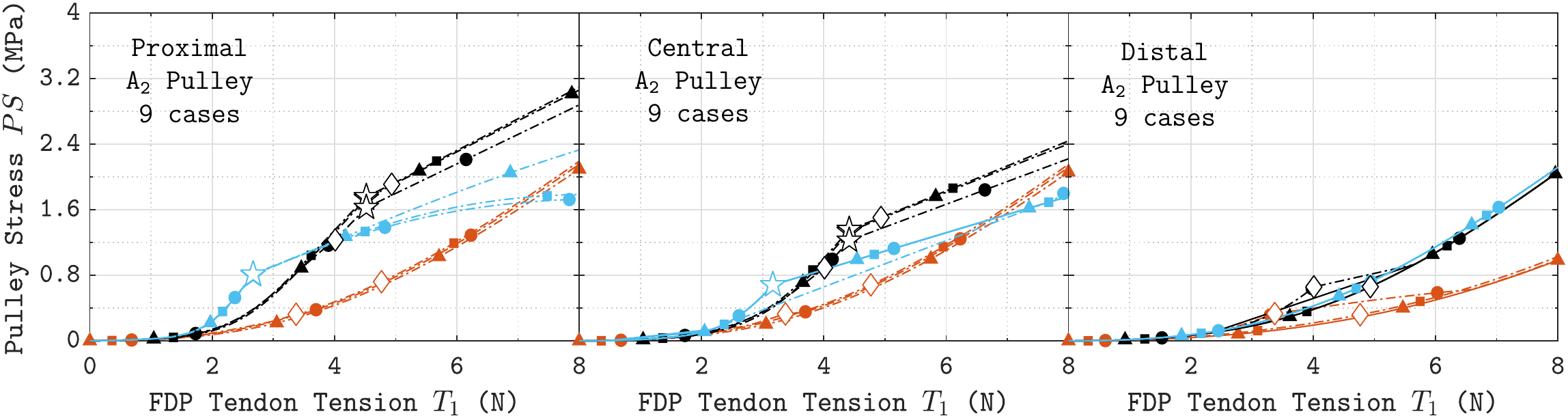} 
	\caption{a-c, left-right: FDP Pulley-Stress response: Curves arranged similar to Fig. \ref{fig:fdp_tension} and correspond to the same parameter values therein. Larger of the resultant stress at the base of A$_2$ and A$_4$ pulleys was chosen as the critical value. The solid curve indicates that higher stress is in A$_2$ pulley. The chained line corresponds to A$_4$ pulley having higher stress. Lower the curve, better is the TPS configuration.
	}  
	\label{fig:fdp_PS} 
	\vspace{1em}   
\end{figure*}

\begin{figure*}[h!]
	\centering	
	\begin{subfigure}[t]{\columnwidth}
		\centering
		\includegraphics[scale=1]{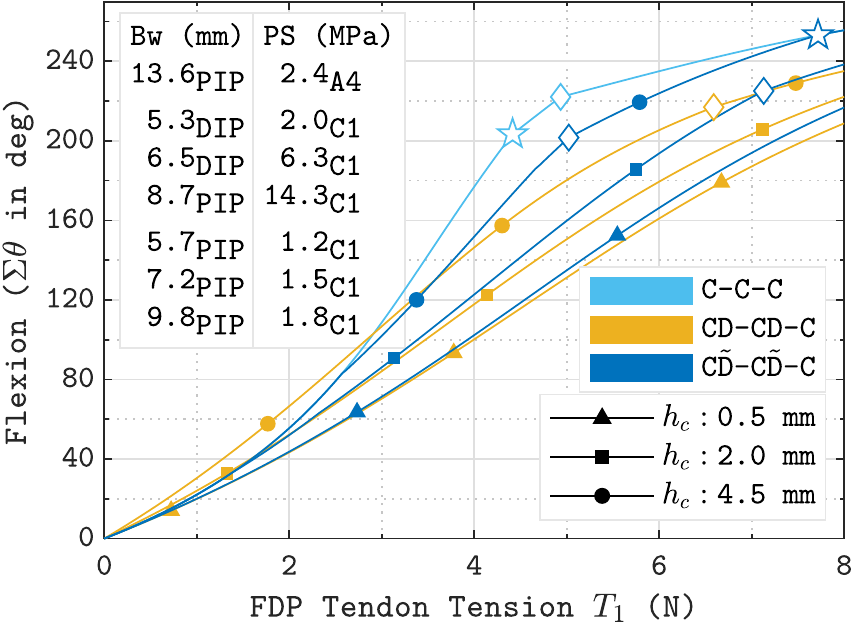}
		\caption{Role of flexible-inelastic C--pulleys and their heights:  $h_a = h_0$. $w_a = w_c = w_0$, $e=e_0$. Higher the height $h_c$ of C--pulleys, higher the ROF, $B_\mathrm{w}$, and $PS$.  Flexible-inelastic C--pulleys yield higher flexion range and lower pulley stress. }	
		\label{fig:fdp_C_loose_height}	
	\end{subfigure}	
	\hfill
	\begin{subfigure}[t]{\columnwidth}
		\centering
		\includegraphics[scale=1]{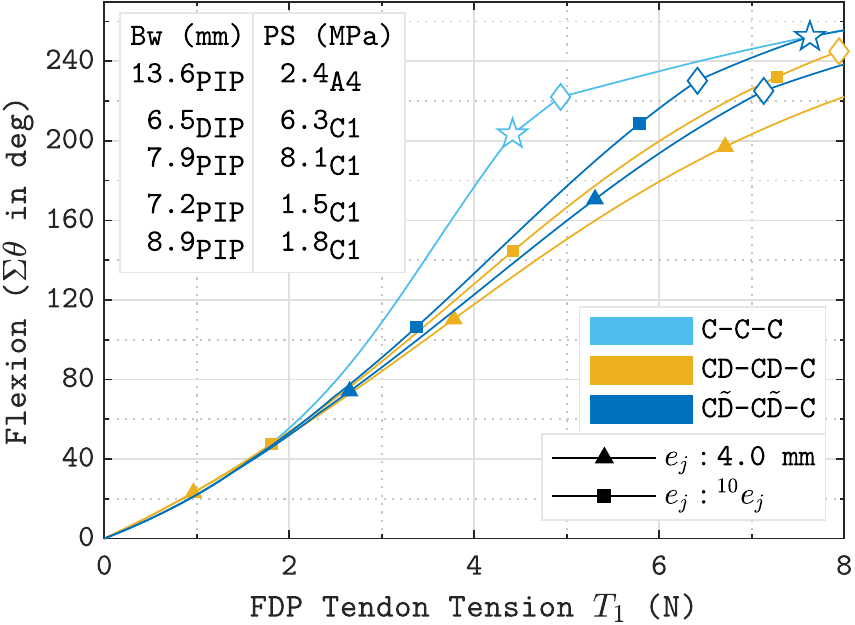}
		\caption{C-Pulley locations:  $h_c = 2.0$ mm, and $h_a = h_0 $. $w_a = w_c = w_0$. ${^{10}e_j} =  (l_j-l_\mathrm{f})/10 + l_\mathrm{f}/2$.   Shifting C-pulleys slightly away from joints results in much higher flexion range, without affecting $B_\mathrm{w}$ and $PS$ much.  }
		\label{fig:fdp_C_location}	
	\end{subfigure}	
	\\[1em]
	\begin{subfigure}[t]{\columnwidth}
		\centering
		\includegraphics[scale=1]{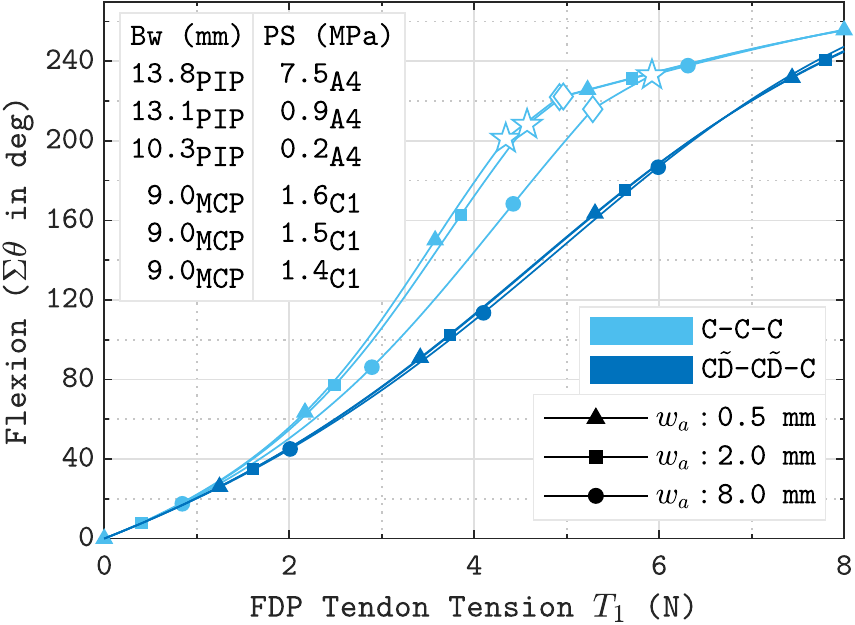}
		\caption{Effect of varying widths $w_a$ of pulleys A$_2$ and A$_4$: $h_c = 2.0$ mm and $h_a = h_5 = h_0$. $e_j = {^{10}e_j}$. $w_c = w_0 $. Higher the widths of A pulleys, lower the pulley stress.}	
		\label{fig:fdp_width}	
	\end{subfigure}	
	\hfill
	\begin{subfigure}[t]{\columnwidth}
		\includegraphics[scale=1]{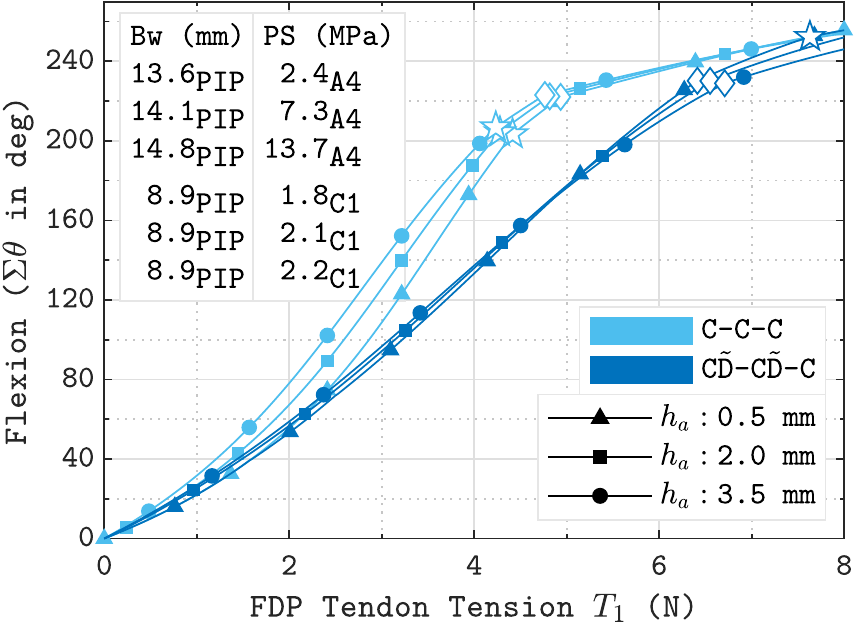}
		\caption{Effect of varying heights $h_a$ of pulleys A$_2$ and A$_4$: $h_c = 2.0$ mm and $h_5 = h_0$. $e_j = {^{10}e_j}$. $w_a = w_c = w_0 $.  Lower the heights of A pulleys, lower the pulley stress.}	
		\label{fig:fdp_height}	
	\end{subfigure}		
	\caption{Effect of parameters and C-pulleys on FDP-TPS: In (a), the first row of values of bowstringing ($B_\mathrm{w}$) and pulley stress ($PS$) is for TPS configuration \tps{C}{C}{C}. The next three values are for the case with stiff C--pulleys. The last three values correspond to the case with flexible-inelastic C--pulleys. For each kind of C--pulley, values are listed in the order of legends corresponding to C--pulley heights $h_c$. Subscripts show locations where these critical values occur. Similar order is followed in (b)-(d) too. In (b), for each kind of C--pulley, values are listed in the order of legends corresponding to C--pulley offsets $e_j$, whereas in (c) to A--pulley heights $h_a$, and in (d) to A--pulley widths $w_a$.  Bowstringing near MCP joint was observed higher (9 mm) in some cases. We excluded it in (a), (b) and (d) to demonstrate the effect of changes in parameters. Star and diamond (hollow) markers indicate PIP and DIP joint locking respectively. }
	\label{fig:fdp_major}
\end{figure*}

\begin{figure}[h!]
	\centering
	\includegraphics[scale=0.79]{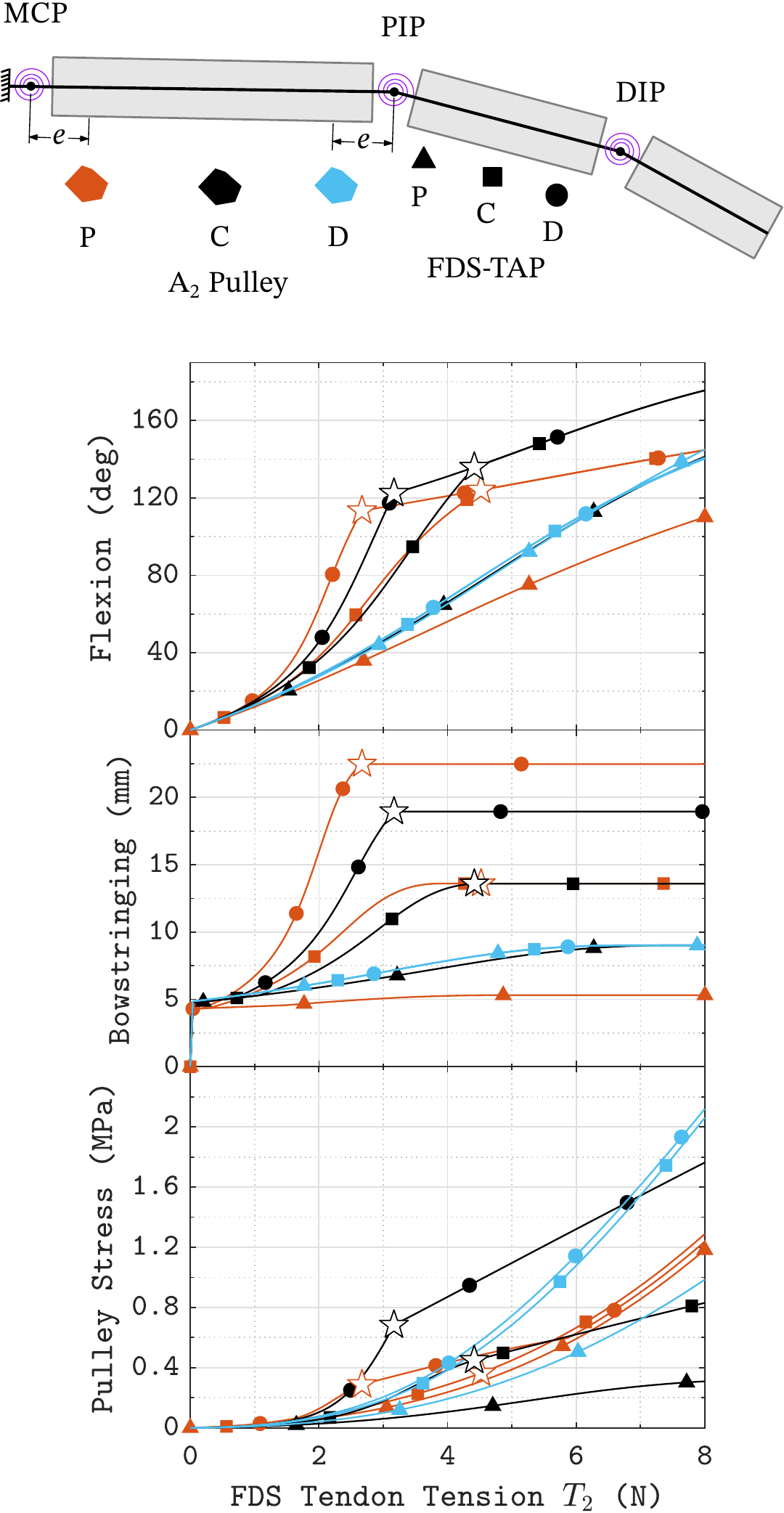}
	\caption{\textcolor{blue}{a-d, top-bottom,} FDS--TPS flexion response: (a) shows the conventions followed in (b), (c), and (d): Three locations --- proximal (P), central (C), and distal (D) considered for A$_2$ pulley (differentiated by colors) and FDS--TAP (differentiated by markers).  Star-marker indicates PIP joint locking. In (c), critical value $B_\mathrm{w}$ of bowstringing is the maximum of MCP and PIP moment arms. The subplot (d) shows critical values of resultant stress at the base of A$_2$ pulley. Pulley/FDS--TAP heights $h_i = h_0$, and widths $w_i = w_0$; $i=1,2 $. In (b), higher the curve, better is the FDP--TPS configuration. In (c) and (d), lower the curve, better is the TPS configuration.}	
	\label{fig:fds_all}
\end{figure}
\begin{figure}[h!] 
	\centering 
	\includegraphics[scale=1]{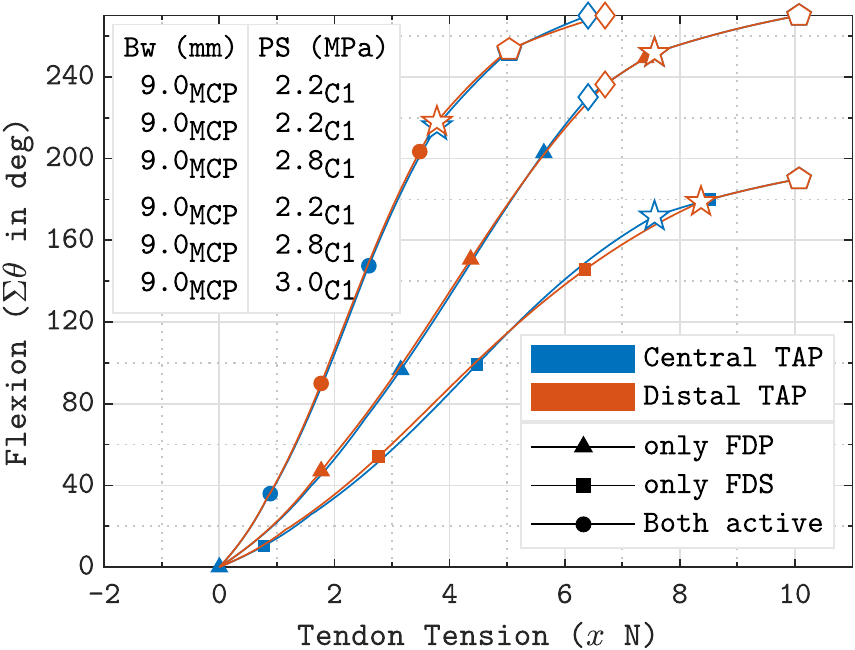}
	\caption{Combined actuation of FDP and FDS tendons in TPS configurations \tps{C\flex{D}}{C\flex{D}}{X},  \tps{C\flex{D}}{X}, and  \fdps{C\flex{D}}{C\flex{D}}{X}, where  FDP/FDS-TAP is central (X = C) or distal (X = D): A$_4$ pulley and FDS--TAP are assumed at the same location in a given TPS configuration.  Delta, square, and disk marked curves correspond to cases when load is exerted by (i) FDP tendon only, (ii) FDS tendon only, and (iii) both tendons together. Pentagon, star, and diamond markers indicate MCP, PIP, and DIP joint locking, respectively.  $h_c = 2$ mm, $w_c = w_0$, and $e_j = {^{10}e_j}$. $w_a = 2$ mm, $h_a = h_0$. The first three values correspond to the case with central FDP/FDS-TAP, while the next three to distal FDP/FDS-TAP. For each case, bowstringing ($B_\mathrm{w}$) and pulley stress ($PS$) values are listed in the order of the legends for markers. Pulley stress is lower for X = C.}     
	\label{fig:fdps}                             
\end{figure}

Using the 3R PRBM, we determined the effect of TPS parameters on the range of flexion (ROF) and critical values of bowstringing $B_\mathrm{w}$ and pulley stresses $PS$. Various finger model data used in the PRBM are listed in Tab. \ref{tab:constants} and results are summarized in Tab. \ref{tab:all_results}.

To explain the observations, we qualified pulleys and TAPs as proximal (P), central (C), or distal (D) based on their locations on the respective phalanges (Fig. \ref{fig:fdp_convention}).  We also named TPS configurations using the characters P, C, and D in groups separated by hyphens (--) or double hyphens (=). The leftmost group describes pulleys on the proximal phalange. The rightmost character describes TAP. For example, \tps{C}{D}{P} indicates one central pulley on the proximal phalange, one distal pulley on the intermediate phalange, and proximal FDP-TAP on the distal phalange. A hyphen (--) indicates that only one tendon is active, whereas double-hyphen (=) indicates that both FDP and FDS tendons are active. In the above example, only FDP tendon is active. Another example \tps{C}{C}{} implying nothing on the distal phalange, indicates FDS tendon, with central pulley on the proximal phalange and central FDS-TAP on the intermediate phalange. \fdps{C\flex{P}}{CD}{C} indicates one central and one proximal pulley on the proximal phalange, and one central and one distal pulley on the intermediate phalange. Both tendons are active. Further, tilde over P indicates that the corresponding pulley is flexible--inextensible.  In each group, the second character is for a C--pulley. The rest are for A--pulleys/TAPs.

To understand the effect of pulley/TAP locations,  we analyzed the case of FDP tendon with one pulley per phalange. We considered 27 TPS configurations, formed from three candidate locations each, of pulleys A$_2$ and A$_4$, and the FDP--TAP, as described in Fig. \ref{fig:fdp_convention}. The pulley heights, widths and offsets used were the default values $h_0$, $w_0$, and $e_0$, respectively, given in Tab. \ref{tab:constants}. In these simulations,  TPS configurations \tps{C}{C}{C} and \tps{C}{C}{D} yield the highest ROF of $256^\circ$ at \SI{8}{\newton} tendon tension, with 13.6 mm bowstringing and pulley stresses of 2.4 and 2.2 MPa, respectively (Fig. \ref{fig:fdp_tension}b--\ref{fig:fdp_PS}b). The pulley A$_4$ experiences higher stress than the pulley A$_2$ (0.8 MPa; Fig. \ref{fig:fdp_PS}b).

To reduce $B_\mathrm{w}$, we added pulleys  C$_1$ and C$_3$ of width $w_0$ at distal locations with offset $e=e_0$ on proximal and intermediate phalanges, respectively. This TPS configuration \tps{CD}{CD}{C} with stiff C--pulleys of 2 mm height yields bowstringing $B_\mathrm{w}$ of 6.5 mm, flexion range of $222^\circ$, and pulley sress 6.3 MPa (Fig. \ref{fig:fdp_C_loose_height}). With C-pulley height of 4.5 mm, bowstringing of 8.7 mm, flexion range of $235^\circ$, and pulley stress of 14.3 MPa are obtained. Replacing stiff C--pulleys by flexible-inelastic ones increases flexion range to  $238^\circ$ with 2 mm C-pulley height, and to  $256^\circ$ with 4.5 mm C-pulley height. Pulley stress remains below $1.8$ MPa for pulley height $\le 4.5$ mm, and bowstringing below 9.8 mm. Positioning the flexible C--pulleys of height $2$ mm at an offset $e_j={^{10}e_j}=(l_j-l_\mathrm{f})/10 + l_\mathrm{f}/2$ from joints (Fig. \ref{fig:fdp_convention}), i.e., 10\% of the respective bone lengths, increases flexion range to $256^\circ$ (Fig. \ref{fig:fdp_C_location}). Bowstringing becomes 8.9 mm. With proximal C--pulleys, as in TPS configuration \tps{C\flex{P}}{C\flex{D}}{C}, flexion range is smaller, and bowstringing and pulley stresses are higher (Tab. \ref{tab:all_results}). 

Without C--pulleys, increasing widths of pulleys A$_2$ and A$_4$ from 0.5 mm (very thin) to 2 mm reduces pulley stress from 7.5 MPa to 0.9 MPa, without affecting flexion range much (Fig. \ref{fig:fdp_width}, Tab. \ref{tab:all_results}).  Increasing heights of pulleys A$_2$ and A$_4$ with width $w_0$, from 0.5 mm to 3.5 mm, increases the pulley stress from 2.4 MPa to 13.7 MPa (Fig. \ref{fig:fdp_height}). Flexion range does not change much, while bowstringing increases from 13.6 mm to 14.8 mm. In presence of flexible C--pulleys, the pulley stress increases from 1.8 MPa to 2.2 MPa, while bowstringing remains unaffected (Fig. \ref{fig:fdp_height}). The flexion range decreases from 256$^\circ$ to 246$^\circ$.

To study the FDS-TPS, we analyzed case of the FDS tendon with one-pulley per phalange. Figure \ref{fig:fds_all}a  describes nine TPS configurations, formed from three candidate positions of A$_2$ pulley and FDS—TAP. As observed in Fig. \ref{fig:fds_all}b-d, TPS configuration \tps{C}{C}{} gives the highest flexion range of $176^\circ$,  bowstringing $B_\mathrm{w}$ of 13.6 mm, and pulley stress $PS$ of 0.8 MPa.  TPS configuration \tps{C}{D}{}  results in higher bowstringing of 18.9 mm, and higher pulley stress of 1.8 MPa. To reduce bowstringing, we added C$_1$ pulley with offset $e_1 = {^{10}e_1}$, width $w_0$, and height 2 mm. With the resulting TPS configurations \tps{C\flex{D}}{C}{} and \tps{C\flex{D}}{D}{},  $B_\mathrm{w}$ reduces to 9.0 mm, and pulley stress increases to 2.2 and 2.8 MPa respectively, without affecting the flexion range (Tab. \ref{tab:all_results}).

To study the combined actuation of FDP and FDS tendons, we arrived at the common configuration \fdps{C\flex{D}}{C\flex{D}}{C}, as follows. For highest flexion range, FDP--TPS configurations are \tps{C\flex{D}}{C\flex{D}}{C} and \tps{C\flex{D}}{C\flex{D}}{D}, and FDS--TPS configurations are \tps{C\flex{D}}{C}{} and \tps{C\flex{D}}{D}{}  (Tab. \ref{tab:all_results}). Both bowstringing and pulley stress are much lower for FDS--TPS configuration \tps{C\flex{D}}{C}{}. Hence, we merged it with FDP-TPS configuration \tps{C\flex{D}}{C\flex{D}}{C}. When both tendons were actuated in this configuration \fdps{C\flex{D}}{C\flex{D}}{C}, sharing equal loads, full finger flexion (270$^\circ$) was achieved at much lower individual tension of 6.4 N (Fig. \ref{fig:fdps}, Tab. \ref{tab:all_results}). Pulley-stress increased to 2.8 MPa from 2.2 MPa for full flexion with only FDP-tendon achieved at 10.1 N. 

Overall, critical bowstringing was observed mostly at MCP or PIP joint, while critical stress at pulley A$_4$ or C$_1$ (Tab. \ref{tab:all_results}).
\section{Discussion}
\label{sec:discussion}

Results for FDP-TPS with one pulley per phalange show that pulley locations for high ROF ($> 240^\circ$, Tab. \ref{tab:all_results}) suffer from high bowstringing and high pulley stress. This problem can be addressed without affecting ROF adversely, (i) by adding flexible-inextensible C--pulleys slightly away from  joints, and (ii) by either increasing width or decreasing height of annular pulleys or both. Including FDS tendon increases ROF further. We quantified bowstringing $\approx 9$ mm as small, a limit obtained as sum of bone semi-width 3.5 mm, pulley height 2 mm, and the clearance 3 mm beyond pulley height. The proposed analysis can help gain insight into the biological TPS, and also in selecting optimal TPS designs for bionic devices, discussed next.


\subsection{The Biological Tendon Pulley System}
\label{sec:discuss_biological}
We observed that the centrally located annular pulleys A$_2$ and A$_4$ result in very high flexion ranges (Tab. \ref{tab:all_results}), thus agreeing with \cite{dy2013flexor} and \cite{chow2014importance}. Role of flexible-inelastic biological C--pulleys is evident from  significant reduction in pulley stress and increase in ROF compared to identical but stiff pulleys (Figs. \ref{fig:fdp_C_loose_height} and \ref{fig:fdp_C_location}). C--pulleys also lower bowstringing considerably. We also observed that a small increase in the widths of A-pulleys decreases pulley stress significantly (Fig. \ref{fig:fdp_width}). However, pulley width does not affect ROF much, thus concurring with  \cite{mitsionis1999feasibility, chow2014importance, leeflang2014role}. To explain why A$_2$ pulley is the widest, we reckon further analysis is necessary.

A slight increase in heights of the annular pulleys increases pulley stress significantly  in the absence of C--pulleys (Fig. \ref{fig:fdp_height}). This result explains why loosening of the main pulleys A$_2$ and A$_4$ during injury is painful when C--pulleys get torn. In this case, high bowstringing is also observed. However, it remains unexplained why ROF reduces in the case of biological TPS.  Biologically, FDS--TAP is immediately proximal to  pulley A$_4$, and therefore nearly central on IP. Further, both FDP and FDS tendons share the same set of pulleys. Both these aspects can be explained from the observations that FDP-TPS configurations \tps{CD}{CD}{C} and FDS-TPS configurations \tps{CD}{C}{} yield the highest ROF, with low bowstringing and pulley-stress (Fig. \ref{fig:fdps}).  

Most of the above aspects indicate that biological TPS has evolved to maximize ROF with minimum possible actuation tension, bowstringing and pulley stress. Pulleys A$_1$, A$_3$, A$_5$ and C$_2$ in Fig. \ref{fig:tendon_pulley}, considered less important in literature, were excluded herein.  

\subsection{Bionic devices based on tendon-pulley system}
Recently, flexure hinges and TPS based robotic hand devices have shown to be useful in rehabilitation and daily assistance of hand-impaired patients (\cite{hofmann2018design, mutlu2015effect}). This study may help improve existing designs by offering the right TPS configuration as per the requirements. To exemplify, consider developing a TPS based hand orthosis with one pulley per phalange and one (FDP) tendon. Results for FDP--TPS at 8 N tendon tension (Figs. \ref{fig:fdp_tension}-\ref{fig:fdp_PS}) show that for small bowstringing ($B_\mathrm{w} \le$ 9 mm), the highest ROF attainable is $210^\circ$ which can be increased to $225^\circ$, if the $B_\mathrm{w}$ limit is increased to 10 mm. Thus, only a suboptimal design can be obtained. The same ROF can be obtained if desired, at a lower tendon tension with two pulley per phalange designs. 
In that case, we recommend the TPS configuration \tps{C\flex{D}}{C\flex{D}}{C} with flexible-inelastic C--pulleys of 2 mm height, offset from joints by 10\% of the respective bone lengths (Fig. \ref{fig:fdp_C_location}). This configuration offers the ROF of $210^\circ$ at 5.5 N tension and the full ROF (270$^\circ$) at 10.1 N tension without compromising on both the bowstringing and pulley stresses. Adding FDS tendon helps achieve the full ROF at a much lower individual tendon tension of 6.4 N (Fig. \ref{fig:fdps}). In this case, two smaller actuators may be used with a control system to distribute the load in both tendons, and thereby generate several finger flexion postures (Fig. \ref{fig:fdps}). However, the design may become bulky and also require a complex control system. A single FDP tendon may be sufficient when designing orthosis for hand open-close exercises, in case forming different hand postures is not essential. In this case, the TPS configuration suggested by \cite{hume1991biomechanics} (explained in section \ref{sec:tps_reconstruction}) is also optimal.     
These examples highlight the importance of the parametric study in choosing an optimal flexor TPS configuration given the design requirements and making aware of the trade-offs needed.

\begin{figure}[h!]
	\centering
	\includegraphics[scale=1]{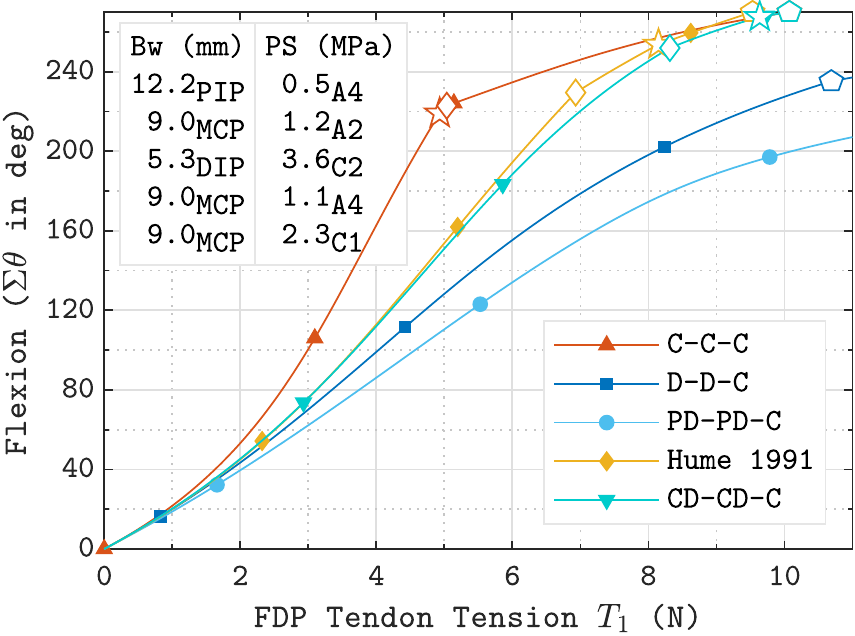}
	\caption{
		FDP--TPS configurations in literature: Case (i) \tps{C}{C}{C}, with $w_a = 4$ mm (\cite{delph2013soft,nycz2015modeling}), Case (ii) \tps{D}{D}{C}, with $w_a = 2$ mm (\cite{bajajsoft}), Case (iii) \tps{PD}{PD}{C}, with $w_a = 2$ mm, $h_c = h_0$, offsets $e_j = e_0$ (\cite{jung2009design,xu2012design,xu2016design}), Case (iv) \tps{PD}{PD}{P} by \cite{hume1991biomechanics}: $w_a = w_c =  2$ mm, $e_k = x_k (l_j - l_\mathrm{f}) + l_\mathrm{f}/2, k=1,2,\cdots,5$, and $j=\lceil k/2 \rceil$, where $\mathbf{x} = (0.21, 0.31, 0.25, 0.21, 0.27)^\mathrm{T}$  (metaphysis dimensions from \cite{SCHULTERELLIS1984490}),  Case (v) \tps{C\flex{D}}{C\flex{D}}{C} (current study) with flexible-inelastic C--pulleys. $e_j = {^{10}e_j}$, $w_a = 2$ mm, $h_c = 2$ mm.    Default values used for heights of A--pulleys ($h_0$), and widths of C--pulleys ($w_0$) in all cases unless mentioned above. Star, and diamond (hollow) markers represent PIP and DIP joint locking respectively. Numerical values are arranged in the order of legends. The parameter values in cases (i)-(iv) are only estimates, and may not be accurate. }    
	\label{fig:tps_literature}                                        
\end{figure}  
We also analyzed various TPS configurations employed in the existing robotic hands (Fig. \ref{fig:tps_literature}) and found them mostly suboptimal. These configurations involve only a single (FDP) tendon. FDP--TPS configuration \tps{C}{C}{C}  used by \cite{delph2013soft} and \cite{nycz2015modeling} suffers from  high bowstringing (3.2 mm above the limit). Configuration \tps{D}{D}{C} with 2 mm pulley width by \cite{bajajsoft} results in small bowstringing, but also a much smaller flexion range (225$^\circ$ at 10 N). Configuration \tps{PD}{PD}{C} by \cite{jung2009design}, \cite{xu2012design}, and \cite{xu2016design} results in even lower flexion range (215$^\circ$) and higher pulley stress.

\subsection{Comparison with TPS Reconstruction Literature}
\label{sec:tps_reconstruction}
The average FDP tendon tension in biological fingers for full flexion is 8.15 N (\cite{yang2016assessing}). When only the FDP tendon is active, nearly full flexion at 8 N FDP tension is observed with FDP-TPS configuration \tps{C\flex{D}}{C\flex{D}}{C} (Fig. \ref{fig:fdps}). This reinforces the argument of reconstructing only the FDP tendon, if just one tendon can be repaired (\cite{kotwal2005neglected}).  
The TPS configuration with two pulleys around each joint at the flare of the metaphysis of the phalangeal bones, of small height as suggested by \cite{hume1991biomechanics}  for reconstruction surgery, is also observed to be good in our simulations (Fig. \ref{fig:tps_literature}). High flexion range, in this case, can be attributed to the fact that the two pulleys on each of the proximal and intermediate phalanges behave like a single, very wide pulley located centrally, resulting in the TPS configuration \tps{C}{C}{C}. The effectively large width also helps in lowering both pulley stress and bowstringing. Recommendations of \cite{kauko1967positioning} to use one pulley per phalange near or on joints result in low ROF, as observed in Fig. \ref{fig:fdp_tension}.

\subsection{Limitations, Advantages, and Future Scope}  
Results herein are based on a specific set of finger joint stiffness values (section \ref{sec:results}), assuming the fully extended finger as its neutral state.  With different sets of stiffnesses and neutral states, the relative positioning of curves corresponding to different TPS configurations is expected to remain similar. Therefore, it may not affect the choice of TPS configuration much. This can be explained geometrically from Fig. \ref{fig:prbm}a based on equilibrium moment-arm variations.  Nevertheless, one may need to verify by regenerating all graphs as per her/his finger joint stiffnesses.  
Some patients having spasticity\footnote{ Spasticity is a condition in which fingers are always in the flexed state and resist extension.} or otherwise, have either joint neutral positions or joint stiffnesses or both altered. An advantage with the 3R PRBM computational model used herein is that it is readily adaptable in such situations.

This study does not address much on the coordination between FDS and FDP tendons. That requires simulating grasping using contact mechanics, which can provide insight into how the two tendons share the load in forming different finger postures. An extensor mechanism can also be included, as it is known to contribute during grasping.

\section{Conclusions}
\label{sec:conclusion}

The presented study facilitates choosing an optimal flexor tendon pulley system (TPS) configuration based on one's design requirements, while also making aware of the trade-offs needed. It also explains several aspects of the biological TPS. This fact validates our study, as well as indicates that the objective of high flexion range with low tendon tension, bowstringing, and pulley stress  is in accordance with nature. We also demonstrated that TPS configurations superior to those used in existing hand prosthetic devices could be employed without introducing much additional complexity. Results herein may alter, but only quantitatively, in presence of hand abnormalities through variations in joint stiffnesses and neutral positions. 

This study may find applications in -- (i) understanding flexion biomechanics of the human finger, (ii) designing cost-effective robotic devices for hand, and (iii) surgery related to tendon pulley reconstruction.

\begin{appendix}
	
\section{Nonlinear Finite Element Model}
\label{app:fem}
\begin{figure}[h!]  
	\centering  
	\includegraphics[scale=1]{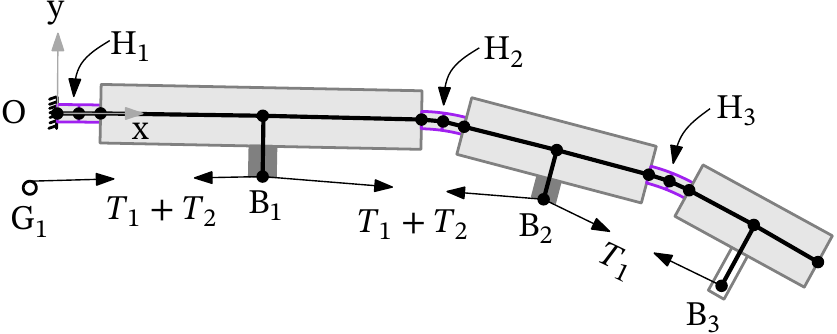}
	\caption{Finite Element Method: Analysed the model with one-pulley per phalange and single (FDP or FDS) tendon during validation, without loss of generality. TPS configuration with pulleys A$_2$ and A$_4$ and FDP/FDS-TAP placed centrally on respective phalanges. Geometry (in gray) discretized using  using 1D co-rotation frame elements. Black disks represent nodes, and thick black line segments depict elements. Two elements per flexure shown as an example. Simulations involve 20 elements per flexure. Frictionless point contact is assumed between tendons and pulleys.}  
	\label{fig:model1_discret}	
\end{figure}

In the nonlinear finite element method (FEM) formulation, 1D co-rotation frame elements were used which can undergo large bending deflection but permit small strain. The small strain was ensured via sufficiently large number of elements per flexure. Discretization of the model geometry and boundary conditions are shown in Fig. \ref{fig:model1_discret}. Detailed formulation for 1D co-rotational frame elements based on the works of \cite{crisfield1993non, belytschko1973non, belytschko1979applications} is given in \cite{mankame2004investigations}. String tensions $T_1$ and $T_1 + T_2$ ($T_1$ in FDP and $T_2$ in FDS tendon) at each boundary node B$_1$, B$_2$, and B$_3$ (Fig. \ref{fig:model1_discret}) are always directed towards the neighbouring nodes lying on the string. This makes the nodal external force vector  $\mathbf{f}_\mathrm{ext} \equiv \mathbf{f}_\mathrm{ext}(T, \mathbf{u})$, i.e., dependent on $\mathbf{u}$ where $\mathbf{u}$ is the nodal displacement vector. To solve nonlinear equilibrium equations using the Newton-Raphson technique, with $\mathbf{g} = \mathbf{f}_\mathrm{int} - \mathbf{f}_\mathrm{ext}$ as the force residual where $\mathbf{f}_\mathrm{int}$ is the nodal internal force vector, the tangent stiffness matrix $\mathbf{K}_\mathrm{t}$ was computed as
\begin{align}
\mathbf{K}_\mathrm{t}=\frac{\partial \mathbf{g}}{\partial \mathbf{u}} = \frac{\partial \mathbf{f}_\mathrm{int}}{\partial \mathbf{u}} - \frac{\partial \mathbf{f}_\mathrm{ext}}{\partial \mathbf{u}} 
\end{align}
One notes that $\frac{\partial \mathbf{f}_\mathrm{ext}}{\partial \mathbf{u}} \neq \mathbf{0}$.]

\section{Validation of the computational model}
\label{app:validation}

\begin{figure}[h!]
	\centering
	\includegraphics[scale=1]{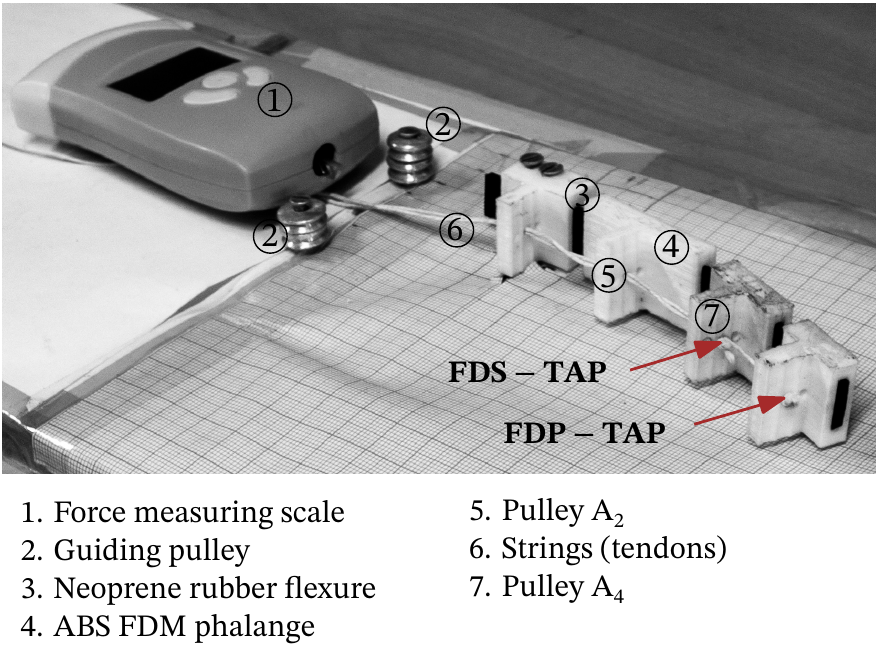}
	\caption{Experimental setup for validation of the computational models: One pulley per phalange model with FDP/FDS tendon in TPS configuration \tps{C}{C}{C} or \tps{C}{C}{}. Neoprene rubber strips used as flexures, which pass through hollow rigid phalanges FDM printed from ABS plastic. Inextensible strings used as tendons. FDP and FDS tendons pulled manually one at a time, to record force-deflection at each load step.}
	\label{fig:TPS_exp_setup}
\end{figure}

To validate the two computational models, we developed a prototype of the biomechanical model (Fig. \ref{fig:TPS_exp_setup}). C-Pulleys were not included. All three phalanges were 3D-printed (FDM) using ABS plastic (2000 MPa, Young's modulus), and had the same cross-section (20 mm $\times$ 6 mm). A neoprene rubber (9 MPa, Young's modulus) strip of cross-section 11.6 mm $\times$ 2.1 mm was used as flexure for finger joints.  7.5 mm height (includes bone-width in the model, Fig. \ref{fig:prbm}) was chosen for all pulleys including the guiding pulley, and the TAPs. The guiding pulley G$_1$ location was chosen to be $(-10, -7.5)$ mm (refer Fig. \ref{fig:model1_discret}). 

To reduce friction between the platform and the prototype, we embedded a 4 mm carbon-steel ball on each phalange. A pull-type force dynamometer with 2 gf resolution was employed to measure the string tension. To account for measurement errors, we conducted five trials of finger flexion for each of the FDP and FDS tendons. Finally, we compared the mean and standard deviation of the tendon tension and flexion range with the simulation results from both FEM and 3R PRBM.

\begin{figure*}[h!]
	\centering
	\begin{subfigure}{\columnwidth}
		\centering
		\includegraphics[scale=1]{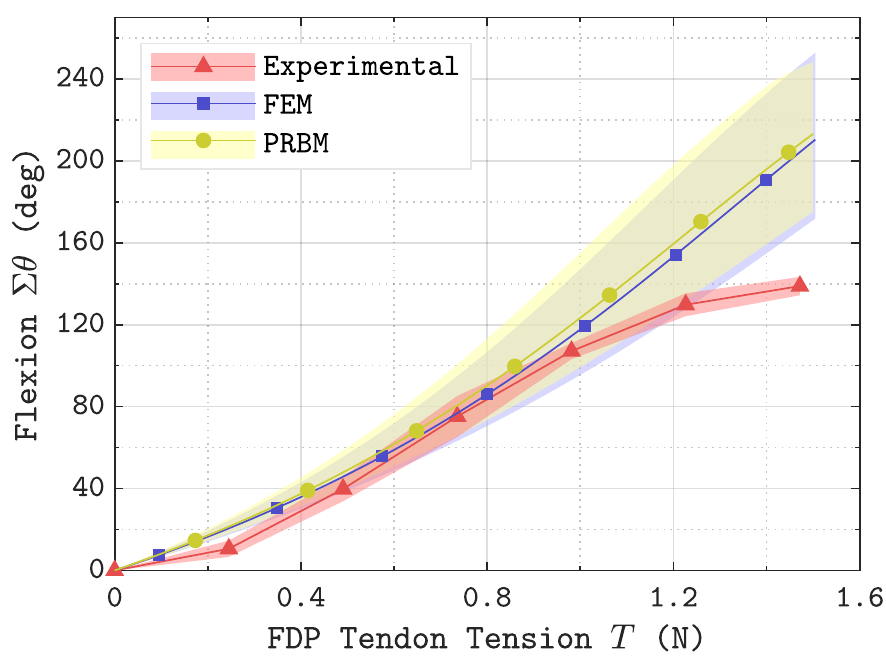}
		\caption{}
		\label{fig:fdp_exp}
	\end{subfigure}
	\hfill
	\begin{subfigure}{\columnwidth}
		\centering
		\includegraphics[scale=1]{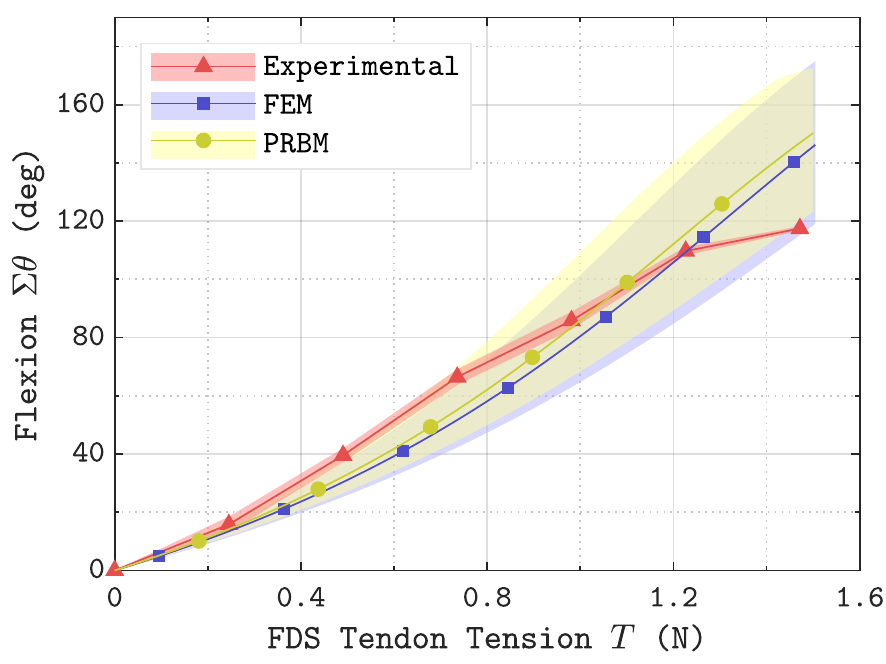}
		\caption{}
		\label{fig:fds_exp}
	\end{subfigure}
	\caption{Comparison of experimental and simulation results for (a) FDP and (b) FDS tendon: Tendons not actuated simultaneously. To account for measurement errors, five trials conducted for each tendon. Standard deviation for the experimental curve shown with the red patch. Rubber strip manufacturing tolerances simulated with the 3R PRBM and FEM, and its results shown with blue and yellow patches respectively. Curves are quite close up to $110^\circ$ of flexion defined by $\sum{\theta_i}$. The experimental curve flattens thereafter, possibly due to material nonlinearity in the neoprene rubber.}
	\label{fig:validation}
\end{figure*}

To account for manufacturing tolerances (available rubber strip thickness = 2.1 $\pm$ 0.1 mm, width = 11.6 $\pm$ 0.1 mm), we simulated the computational model for flexure dimensions in this tolerance band. The strip width was chosen equal to the finger width. The strip thickness was fixed to ensure that flexure stiffness matches that of the respective finger joint. The moment exerted on the right portion of the finger (FBD in Fig. \ref{fig:prbm}b) can be obtained from FEM as the moment at its leftmost node. This moment should be equal to that obtained from PRBM. With this understanding, we performed some trial and error with the strip thickness for each flexure to arrive at its appropriate thickness. Equivalent stiffness of all three joints with the nominal dimensions is $\approx \SI{0.27}{\newton \milli\meter}$. The upper tolerance value corresponds to $\approx\SI{0.31}{\newton \milli\meter}$, whereas the lower one to $\approx\SI{0.23}{\newton \milli\meter}$. PRBM is used to simulate the TPS configuration \tps{C}{C}{C} or \tps{C}{C}{} for these three stiffness sets. FEM directly uses the highest and lowest dimensions in addition to the nominal dimensions. As a result, we obtained the bands of tension-flexion response as shown in Fig. \ref{fig:validation}.

All three methods yield very similar results in the limit of standard deviation, up to $\Sigma\theta = 120^\circ$, at tendon tension of 1.2 N for FDP tendon. After that, the experimental curve diverges, which may be explained by the material nonlinearity associated with the neoprene rubber corresponding to large deflections. Both FEM and PRBM as implemented herein, disregard material nonlinearity.  Results for the FDS tendon are similar (Fig. \ref{fig:fds_exp}).

\end{appendix}

\bibliography{references}

\end{document}